\documentclass[default,iicol]{sn-jnl}%

\usepackage{caption}
\usepackage{romannum}
\usepackage{multicol,lipsum}

\jyear{2021}%

\theoremstyle{thmstyleone}%

\theoremstyle{thmstyletwo}%

\theoremstyle{thmstylethree}%

\newcommand{\update}[1]{\textcolor{black}{#1}}

\begin{document}

\title[Article Title]{Multi-Robot Geometric Task-and-Motion Planning for Collaborative Manipulation Tasks}

\author*[1]{\fnm{Hejia} \sur{Zhang}}\email{hejiazha@usc.edu}

\author[1]{\fnm{Shao-Hung} \sur{Chan}}\email{shaohung@usc.edu}

\author[1]{\fnm{Jie} \sur{Zhong}}\email{jzhong54@usc.edu}

\author[2]{\fnm{Jiaoyang} \sur{Li}}\email{jiaoyangli@cmu.edu}

\author[3]{\fnm{Peter} \sur{Kolapo}}\email{peter.kolapo@uky.edu}

\author[1]{\fnm{Sven} \sur{Koenig}}\email{skoenig@usc.edu}

\author[3]{\fnm{Zach} \sur{Agioutantis}}\email{zach.agioutantis@uky.edu}

\author[3]{\fnm{Steven} \sur{Schafrik}}\email{steven.schafrik@uky.edu}

\author[1]{\fnm{Stefanos} \sur{Nikolaidis}}\email{nikolaid@usc.edu}

\affil*[1]{\orgdiv{Department of Computer Science}, \orgname{University of Southern California}, \orgaddress{\country{USA}}}

\affil[2]{\orgdiv{The Robotics Institute}, \orgname{Carnegie Mellon University}, \orgaddress{\country{USA}}}

\affil[3]{\orgdiv{Department of Mining Engineering}, \orgname{University of Kentucky}, \orgaddress{\country{USA}}}

\abstract{We address multi-robot geometric task-and-motion planning (MR-GTAMP) problems in \textit{synchronous}, \textit{monotone} setups. The goal of the MR-GTAMP problem is to move objects with multiple robots to goal regions in the presence of other movable objects. We focus on collaborative manipulation tasks where the robots have to adopt intelligent collaboration strategies to be successful and effective, i.e., decide which robot should move which objects to which positions, and perform collaborative actions, such as handovers. To endow robots with these collaboration capabilities, we propose to first collect occlusion and reachability information for each robot by calling motion-planning algorithms. We then propose a method that uses the collected information to build a graph structure which captures the precedence of the manipulations of different objects and supports the implementation of a mixed-integer program to guide the search for highly effective collaborative task-and-motion plans. The search process for collaborative task-and-motion plans is based on a Monte-Carlo Tree Search (MCTS) exploration strategy to achieve exploration-exploitation balance. We evaluate our framework in two challenging MR-GTAMP domains and show that it outperforms two state-of-the-art baselines with respect to the planning time, the resulting plan length and the number of objects moved. We also show that our framework can be applied to underground mining operations where a robotic arm needs to coordinate with an autonomous roof bolter. We demonstrate plan execution in two roof-bolting scenarios both in simulation and on robots.}


\maketitle

\begin{figure*}[!t]
\centering

\includegraphics[width=0.75\linewidth]{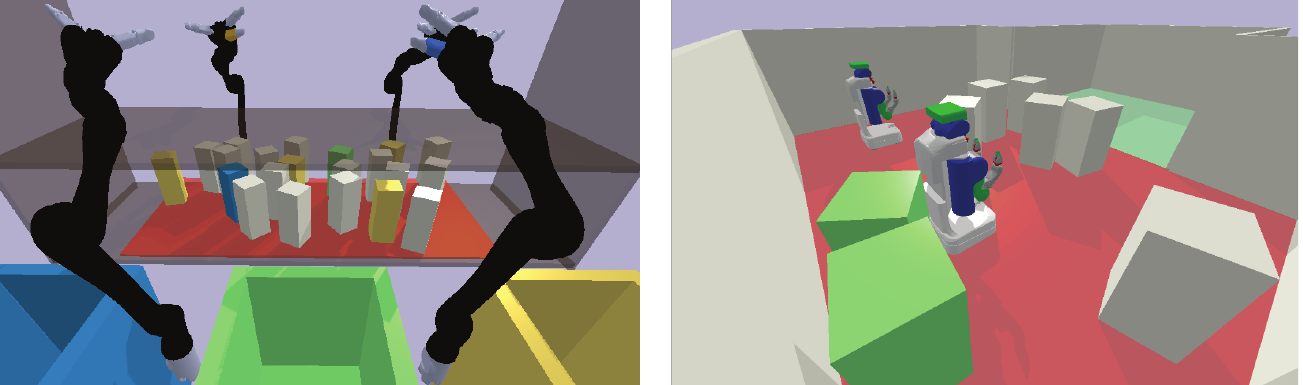}
\caption{Left: \update{Sorting colored objects into boxes of corresponding colors}. Right: Moving the colored boxes to the green region. \update{In both scenarios, white objects are only allowed to be relocated within their current regions (red).} We use PyBullet~\cite{coumans2019} as our simulator.}
\label{fig:best}
\end{figure*}

\section{Introduction}\label{sec1}

Task-and-motion planning (TAMP) is the problem of combining task and motion planning to divide an objective, such as assembling a table, into a series of robot-executable motion trajectories~\cite{doi:10.1146/annurev-control-091420-084139}. Task planning is used to generate a sequence of discrete actions, such as picking up a screwdriver and driving a screw, while motion planning is used to compute the actual trajectories the robot should execute.

Geometric task-and-motion planning (GTAMP) is an important subclass of TAMP where the robot has to move several objects to regions in the presence of other movable objects~\cite{Kim2019}. GTAMP has been addressed efficiently in single-robot domains~\cite{Kim2019,pmlr-v100-kim20a,doi:10.1177/02783649211038280}. We focus on \textit{multi-robot geometric task-and-motion planning} (MR-GTAMP), where several robots have to collaboratively move several objects to regions in the presence of other movable obstacles.

MR-GTAMP naturally arises in many multi-robot manipulation domains, such as multi-robot construction, multi-robot assembly and autonomous warehousing~\cite{chen2022cooperative,hartmann2021long}. MR-GTAMP is interesting as multi-robot systems can perform manipulation tasks more effectively than single-robot systems and can also perform manipulation tasks that are beyond the capabilities of single-robot systems~\cite{9357998}. For example, in a product-packaging task, a single robot may have to move a lot of objects to clear a path to grasp an object, while a two-robot system can easily perform a handover action to increase the effectiveness of task execution. 

\update{Examples of MR-GTAMP problem instances are shown in Fig.~\ref{fig:best}. The example task shown in Fig.~\ref{fig:best} (left) requires multiple robotic arms to sort colored objects into boxes of corresponding colors in a confined workspace. The example task shown in Fig.~\ref{fig:best} (right) requires multiple mobile manipulators to move green objects to the green region. In both tasks, white objects are movable obstacles and are only allowed to be relocated within their current regions. These example tasks embody the key challenges that MR-GTAMP aims to address. First, they are in a hybrid discrete-continuous planning space which is extremely large when multiple robots are involved~\cite{9636119,doi:10.1177/02783649211038280}. This involves high-level task planning, which decides which robot should move which objects and in what sequence, and low-level motion planning, which decides the positions to which objects should be relocated and the motion trajectories robots should follow. Second, in both scenarios, robots work in a confined workspace and have to consider geometric constraints imposed by the environments and the tasks carefully. Finally, robots must collaborate intelligently to perform tasks effectively. For example, robots can achieve their targets more quickly by concurrently manipulating multiple objects, and they can avoid relocating too many objects by performing handover actions.}

We address the following research question: How can we \update{enable} multiple robots to perform GTAMP tasks effectively and efficiently?

Determining effective collaborative action sequences for multiple robots is difficult as manipulation planning in the presence of movable obstacles has been shown to be NP-hard for single-robots~\cite{4209604,9197485}. MR-GTAMP is even harder since one needs to decide which robot should move which objects to which positions.

Our key insight to solving MR-GTAMP efficiently is that \textit{we can compute information about the manipulation capabilities of individual robots and their potential collaborative relationships by calling motion-planning algorithms} and then use it to prune the search space and guide the search process. For example, based on the information that a robot cannot reach an object, we can eliminate all task plans that involve the action where the robot has to reach the object. Moreover, the computed information can be used to generate collaborative plans where each robot performs the tasks that it excels at.

\begin{figure*}[!t]
\centering
\includegraphics[width=1\linewidth]{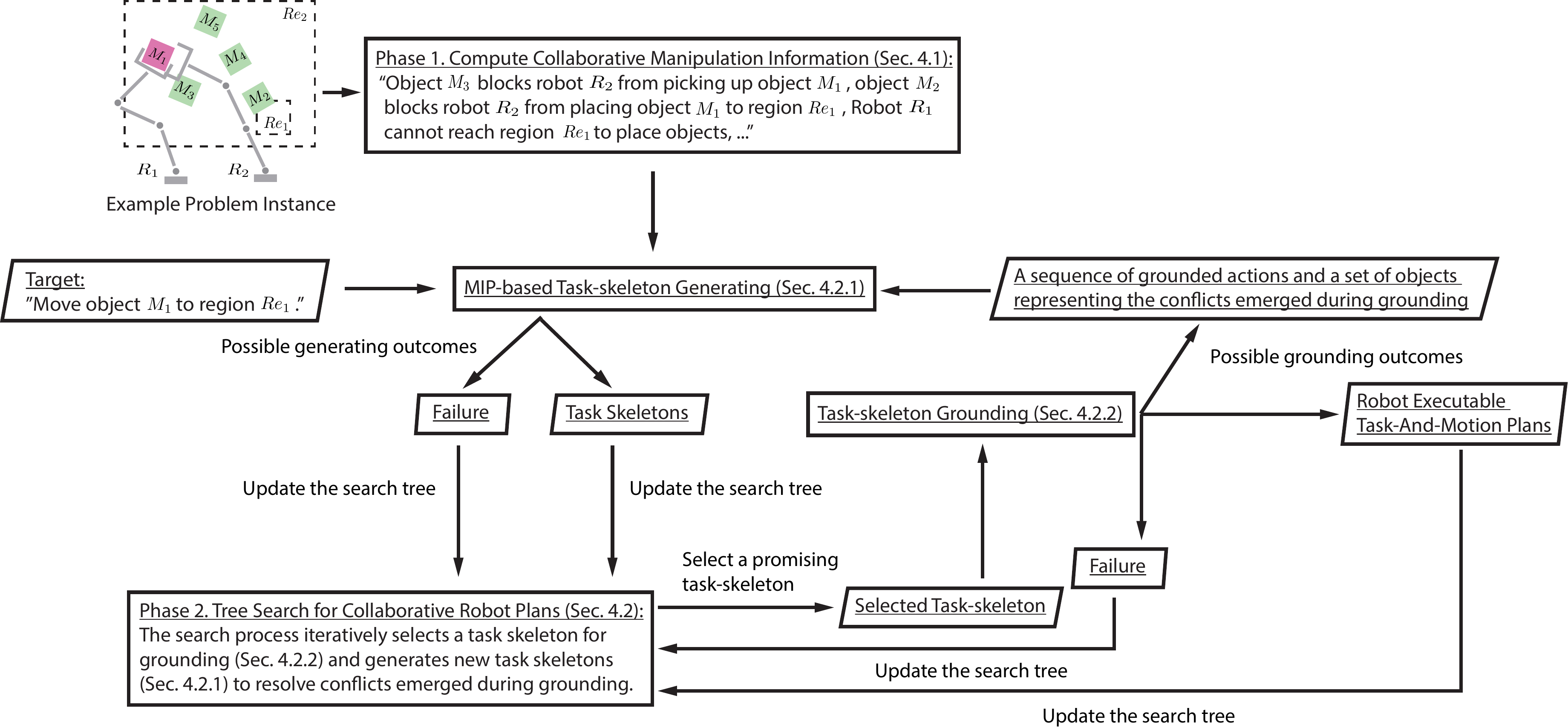}
\caption{Overview of the proposed framework. \update{Fig.~\ref{fig:tree_search} provides a more detailed visualization and description of Phase 2.}}
\label{fig:summary}
\end{figure*}

We propose a two-phase framework. In the first phase, we compute the collaborative manipulation information, i.e., the occlusion and reachability information for individual robots and the potential collaborative relationships between them (Sec.~\ref{sec:rep}). In the second phase, we search for collaborative task-and-motion plans using a Monte-Carlo Tree Search (MCTS) exploration strategy due to its good exploration-exploitation balance (Sec.~\ref{sec:tree_search}). \update{Our search algorithm is based on two key components: (\romannum{1}) The first key component uses the collected information from the first phase to generate promising task skeletons for moving a specified set of objects by formulating a series of mixed-integer linear programs (MIPs), that can be solved efficiently by leveraging recent developments in MIP solvers~\cite{cplex2009v12} (Sec.~\ref{sec:approx}). The term \textit{task skeleton} represents a sequence of actions that are missing continuous parameters required for execution. The missing continuous parameters include the intended positions for objects that need to be relocated, and the motion trajectories that the robots should follow to relocate these objects. The formal definition of \textit{task skeleton} can be found in Sec.~\ref{sec:formulation}. (\romannum{2}) The second key component efficiently finds feasible continuous parameters for the generated task skeletons, such as the locations to which to relocate objects (Sec.~\ref{sec:inst}). We denote the process of finding continuous parameters to make a task skeleton executable as \textit{grounding}.} Fig.~\ref{fig:summary} presents an overview of our framework.

We compare our framework with two state-of-the-art baselines, namely, a general MR-TAMP framework~\cite{9636119} and a multi-robot extension of the ResolveSpatialConstraints (RSC) algorithm~\cite{4209604}. We evaluate our framework in two challenging MR-GTAMP domains and show that it outperforms two state-of-the-art baselines with respect to the planning time, the resulting plan length and the number of objects moved (Sec.~\ref{sec:exp}).

We also conducted an application study and show that our framework can be used to coordinate a robotic arm with an autonomous roof bolter for underground mining operations. We demonstrate the execution of the computed plans in two example roof-bolting scenarios both in simulation and on robots.

Our work makes the following assumptions, which are common in MR-TAMP~\cite{9357998,9636119}: \update{(\romannum{1}) It considers only \textit{monotone} instances of the MR-GTAMP problem, where each object is moved only once. The monotone problems are common in less constrained environments such as home environments and relate to a range of warehouse applications such as packing and stowing~\cite{9357998}. (\romannum{2}) It assumes } the robots \textit{synchronously} start and stop the executions of actions. We plan to relax these assumptions in future work.

This work is an extended version of our prior paper~\cite{zhang2022mip}. We make the following additional contributions.

\begin{itemize}
    \item We conduct an application study on the roof-bolting task, which is an essential operation within the underground mining cycle. We show that the roof-bolting task can be formulated as MR-GTAMP problems and addressed efficiently with the proposed planning framework. We demonstrate plan execution in two roof-bolting scenarios both in simulation and on real robots.
    \item We conduct additional scalability evaluation experiments to study the performance change of our framework when more robots are involved.
    \item We substantially expand the description of the task-skeleton grounding component and the tree search algorithm.
\end{itemize}

\section{Related Work}\label{sec2}

\update{There has been much work on solving general TAMP problems efficiently. TAMP problems are challenging because they require search in a large hybrid space that consists of task-level search and motion-level search. Different approaches for TAMP problems focus on different strategies to combine task-level search and motion-level search. In~\cite{lagriffoul2014efficiently,bidot2017geometric}, efficient geometric backtracking algorithms are proposed to systematically consider all the combinations of geometric instances of a given symbolic task plan such that the symbolic task plan can be efficiently rejected if there is no way to instantiate it geometrically. In~\cite{dantam2018incremental}, task-level search is modeled as a constraint satisfaction problem and failures on motion-level search are efficiently encoded as new constraints to inform task-level search. In~\cite{srivastava2014combined}, an extensible planner-independent interface layer is proposed to combine task and motion planning. In~\cite{garrett2020pddlstream}, motion-level facts are encoded in task-level planning and modern task planners~\cite{Hoffmann2001FFTF} are leveraged to efficiently search for task-and-motion plans. Recently, more and more work has been focused on utilizing learning to guide TAMP by ranking task plans~\cite{10035989}, predicting feasibility of task plans~\cite{yang2022sequence}, and ranking object importance in problem instances~\cite{silver2020planning}. More comprehensive surveys on TAMP can be found in~\cite{doi:10.1146/annurev-control-091420-084139,10.3389/frobt.2021.637888}.}

\update{In this work, we focus on GTAMP which is an important subclass of TAMP. The goal of the GTAMP is to move several objects to regions in the presence of other movable objects. 
}

There has been much work on solving single-robot GTAMP (SR-GTAMP) problems efficiently~\cite{doi:10.1177/02783649211038280,Kim2019,pmlr-v100-kim20a} by utilizing learning to guide planning. \update{However, these approaches cannot be directly applied to multi-robot domains.} Several problem types in the literature can also be seen as versions of the GTAMP problem. In~\cite{4209604}, the ``manipulation among movable obstacles" (MAMO) problem is addressed, in which a robot has to move objects out of the way to move a specified object to its goal location. \update{Although this approach can be extended to multi-robot settings straightforwardly, it would require searching through a large space of all possible combinations of multi-agent actions. Moreover, the focus of this approach is on feasibility of the task-and-motion plans, rather than on the plan length and number of objects moved.} In~\cite{9197485} and~\cite{9196652,8794143}, the object retrieval problem is addressed, in which a robot has to retrieve a target object from clutter by relocating the surrounding objects. In~\cite{7487583,7487581}, the rearrangement planning problem is addressed, in which a robot has to move objects into given goal configurations. However, these methods do not plan collaboration strategies in multi-robot domains.

There has been work on solving general TAMP with several robots efficiently~\cite{9636119,7989464,10.3389/frobt.2021.637888}. We focus on a subclass of these problems, where a robot has to move objects in the presence of movable obstacles. \update{In~\cite{9636119}, a novel task scheduling layer, positioned between task planning and motion planning, is proposed to prune task planning search space. However, since this approach does not focus on geometric aspects of the TAMP problem, it does not include guidance for finding continuous parameters such as feasible positions for object relocation.} In~\cite{7759624,ahn2021coordination}, efficient approaches are proposed for the multi-robot object retrieval problem, assuming permanent object removal and considering one target object at a time, while our planner relocates the obstacles within the workspace and considers several target objects at the same time. Multi-robot rearrangement planning problems~\cite{9357998,hartmann2021long,chen2022cooperative} are also closely related to MR-GTAMP. However, the rearrangement planning problems assume that the goal configurations of all the movable objects are given, while MR-GTAMP requires the planners to decide which objects to move and to which positions. There is also work that focuses on task allocation and scheduling for multiple robots, assuming that a sequence of discrete actions to be executed is given~\cite{9197103}. However, MR-GTAMP requires the planners to decide which discrete actions to execute, e.g., which objects to move.

There has been work on optimization-based TAMP, where TAMP problems are modeled as mixed-integer non-linear programs~\citep{10.5555/2832415.2832517,toussaint2017multi}, mixed-integer linear programs~\citep{kogo2021fast} and continuous nonlinear programs~\citep{9561209}. However, these frameworks do not focus on scenarios where obstacle avoidance is the major challenge and objects can be moved to enable the manipulation of other objects.

\section{Problem Formulation}\label{sec:formulation}

In an MR-GTAMP problem, we have a set of $n_\mathbf{R}$ robots $\mathbf{R} = \{R_i\}^{n_{\mathbf{R}}}_{i=1}$, a set of fixed rigid objects $\mathbf{F}$, a set of $n_\mathbf{M}$ movable rigid objects $\mathbf{M} = \{M_i\}^{n_\mathbf{M}}_{i=1}$ and a set of $n_\mathbf{Re}$ regions $\mathbf{Re} = \{{Re}_i\}^{n_{\mathbf{Re}}}_{i=1}$. We assume that all objects and regions have known and fixed shapes. The focus of our work is not on grasp planning~\cite{7743537}. So, for simplicity, we assume a fixed set of grasps $\mathbf{Gr}_{M, R}$ for each object $M \in \mathbf{M}$ and robot $R \in \mathbf{R}$ pair. $\mathbf{Gr}$ is the union of the sets of grasps for all object and robot pairs.

Each object has a configuration, which includes its position and orientation. Each robot has a configuration defined in its base pose space and joint space. We are given the initial configurations of all robots, objects and a goal specification $\mathcal{G}$ in form of a conjunction of statements of the form $\textsc{InRegion}(M, Re)$, which is true \textit{iff} object $M \in \mathbf{M}$ is contained entirely in region $Re \in \mathbf{Re}$. \update{An example goal specification is $(\textsc{InRegion}(M_1, Re_1) \land \textsc{InRegion}(M_2, Re_1))$ which indicates the target that we want to move objects $M_1$ and $M_2$ to region $Re_1$.}

We define a grounded joint action as a set of $n_{\mathbf{R}}$ actions and motions performed by all the robots at one time step, i.e., the grounded joint action at time step $j$ is an $n_\mathbf{R}$-tuple $s_j = \langle (a_{R_1}^{j}, \xi_{R_1}^{j}), (a_{R_2}^j, \xi_{R_2}^{j}), \dots, (a^j_{R_{n_{\mathbf{R}}}}, \xi_{R_{n_{\mathbf{R}}}}^{j}) \rangle$, where each action $a$ is a pick-and-place action or a wait\footnote{As in~\cite{9636119}, a robot with a wait action does not have to do anything but can move to avoid other robots.} action that the corresponding robot executes and motion $\xi$ is a trajectory that the corresponding robot executes, specified as a sequence of robot configurations. In this work, we focus on pick-and-place actions because of their importance in robotic manipulation in cluttered space. Each pick-and-place action is a tuple of the form $\langle M, Re, R^{pick}, R^{place}, g^{pick}, g^{place}, P_M^{place} \rangle$, where $M$ represents the object to move; $Re$ represents the target region for $M$; $R^{pick}$ and $R^{place}$ represent the robots that pick and place $M$, respectively; $g^{pick}$ and $g^{place}$ represent the grasps used by $R^{pick}$ and $R^{place}$, respectively, and $P_M^{place}$ represents the configuration at which to place $M$.  Moreover, we call a pick-and-place action whose $R^{pick}$ is different from $R^{place}$ a handover action. Each grounded joint action maps the configurations of the movable objects to new configurations and the unaffected objects remain at their old configurations.

We define a partially grounded joint action as an $n_{\mathbf{R}}$-tuple of the form $\langle \bar{a}_{R_1}, \dots, \bar{a}_{R_{n_{\mathbf{R}}}} \rangle$, where $\bar{a}$ is a wait action or a pick-and-place action without the placement information $P_M^{place}$. We refer to a pick-and-place action without the placement information as a partially grounded pick-and-place action since it has only the information about the grasps that will be used.

We define a task skeleton $\bar{\mathbf{S}}$ as a sequence of partially grounded joint actions. We want to find a task-and-motion plan, i.e., a sequence of grounded joint actions $\mathbf{S}$ that changes the configurations of the objects to satisfy $\mathcal{G}$.

\update{We denote the process of finding feasible object placements and motion trajectories for a task skeleton as \textit{grounding}.}

A task-and-motion plan is valid \textit{iff}, at each time step $j$: (\romannum{1}) the corresponding multi-robot trajectory $\Xi^j = \langle \xi_{R_1}^{j}, \xi_{R_2}^{j}, \dots, \xi_{R_{n_{\mathbf{R}}}}^{j} \rangle$ is collision-free; (\romannum{2}) the robots can use the corresponding motion trajectories and grasp poses to grasp the target objects and place them at their target configurations without collisions; and (\romannum{3}) all handover actions can be performed without inducing collisions. The considered collisions include collisions between robots, collisions between an object and a robot and collisions between objects.

\section{Our Approach}\label{sec4}

We present our two-phase MR-GTAMP framework (Fig.~\ref{fig:summary}) in this section. In the first phase, we compute the collaborative manipulation information, i.e., the occlusion and reachability information for individual robots and whether two robots can perform a handover action for an object (Sec.~\ref{sec:rep}). In the second phase, we use a Monte-Carlo Tree Search exploration strategy to search for task-and-motion plans (Sec.~\ref{sec:tree_search}). The search process depends on a key component that generates promising task skeletons (Sec.~\ref{sec:approx}) and a key component that finds collision-free object placements and trajectories for the task skeletons to construct valid task-and-motion plans (Sec.~\ref{sec:inst}).

\subsection{Computing Collaborative Manipulation Information}\label{sec:rep}

Given an MR-GTAMP problem instance and the initial configurations of all objects and robots, our framework first computes the occlusion and reachability information for individual robots, e.g., whether an object blocks a robot from manipulating another object and whether a robot can reach a region to place an object there. We also compute whether two robots can perform a handover action for an object by computing whether they can both reach a predefined handover point to transfer the object. In this work, we consider only handover actions for objects that are named in goal specification $\mathcal{G}$ for computational simplicity. We assume that all robots return to their initial configurations after each time step. Inspired by~\cite{doi:10.1177/02783649211038280}, we use the conjunction of all true instances of a set of predicates to represent the computed information. To define these predicates, we define two volumes of the workspace similar to~\cite{4209604,doi:10.1177/02783649211038280}. The first volume $V_{pick}(M, g, R, \xi)$ is the volume swept by robot $R$ to grasp object $M$ with grasp $g$ following trajectory $\xi$. The second volume $V_{place}(M, g, R, P^{place}_M, \xi)$ is the volume swept by robot $R$ and object $M$ to transfer the object to configuration $P^{place}_M$ following trajectory $\xi$. Our predicates are as follows:

{
\begin{itemize}
    \item \textsc{OccludesPick}$(M_1, M_2, g, R)$ is true \textit{iff} object $M_1$  overlaps with the swept volume $V_{pick}(M_2, g, R, \xi)$, where $\xi$ is chosen to be collision-free with all the objects except $M_2$, if possible;
    \item \textsc{OccludesGoalPlace}$(M_1, M_2, Re, g, R)$ is true \textit{iff} $M_1$ is an object that overlaps with the swept volume $V_{place}(M_2, g, R, P_{M_2}^{place}, \xi)$, where $P_{M_2}^{place}$ and $\xi$ are chosen to be collision-free with all the objects except $M_2$, if possible, and the pair $\langle M_2, Re \rangle$ is named in goal specification $\mathcal{G}$;
    \item \textsc{ReachablePick}$(M, g, R)$ is true \textit{iff} there exists a trajectory for robot $R$ to pick object $M$ with grasp $g$;
    \item \textsc{ReachablePlace}$(M, Re, g, R)$ is true \textit{iff} there exists a trajectory for robot $R$ to place object $M$ into region $Re$ with grasp $g$; and 
    \item\textsc{EnableGoalHandover}$(M, g_1, g_2, R_1, R_2)$ is true \textit{iff} robots $R_1$ and $R_2$ can both reach a predefined handover point for object $M$ with grasps $g_1$ and $g_2$, respectively, and the object $M$ is named in goal specification $\mathcal{G}$. 
\end{itemize}
}

For a predicate instance to be true, the corresponding trajectories are required to be collision-free with respect to all fixed objects. For a predicate instance of \textsc{EnableGoalHandover} to be true, the two robots are required to not collide with each other.

The values of all predicate instances can be computed with existing inverse-kinematics solvers~\cite{diankov_thesis} and motion planners~\cite{lavalle2006planning}. Ideally, we wish to find trajectories for the robots that have the minimum number of collisions with all objects, i.e., the \textit{minimum constraint removal}~\cite{Hauser-RSS-13} trajectories. However, this is known to be very time consuming. Thus, we follow previous work~\cite{doi:10.1177/02783649211038280} and first attempt to find a collision-free trajectory with respect to all movable and fixed objects. If we fail, we attempt to find a collision-free trajectory with respect to only the fixed objects. 

In our implementation, we efficiently compute the predicates  -- with the exception of \textsc{EnableGoalHandover} -- in parallel for all robots by creating an identical simulation environment for each robot.

\begin{figure*}[!t]
\centering
\includegraphics[width=1\linewidth]{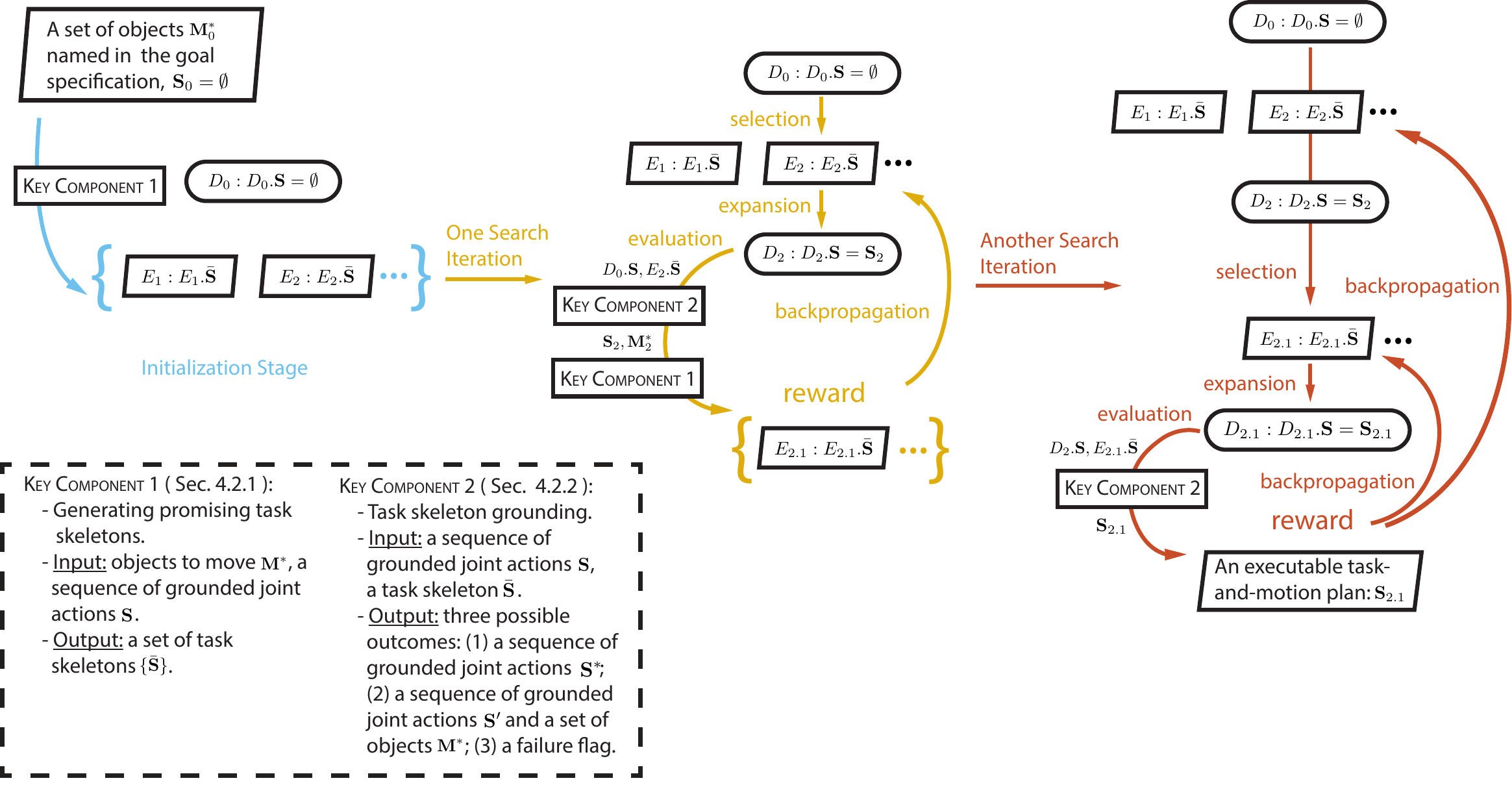}
\caption{\update{Visualization} of the search process in the second phase of our framework. \update{We show the initialization stage of the search process (left) and two example search iterations (middle, right) that lead to different evaluation outcomes.} Left: Blue arrows represent the workflow for initializing the search tree. Middle: Yellow arrows represent a search iteration that results in an updated set of objects to be moved and thus a new set of task skeletons to be grounded. Right: Red arrows represent a search iteration that results in an executable \update{task-and-motion} plan.}
\label{fig:tree_search}
\end{figure*}

\subsection{Searching for Task-and-Motion Plans}\label{sec:tree_search}

We now describe our search process (Fig.~\ref{fig:tree_search}) for efficiently finding effective collaborative task-and-motion plans. \update{Our search process is initialized with a set of task skeletons, that is generated for moving the set of objects named in the goal specification, utilizing the computed collaborative manipulation information (Sec.~\ref{sec:rep}). We will describe our key component for generating task skeletons in detail in Sec.~\ref{sec:approx}. We then generate a search tree with a root node, denoted as $D_0$ as shown in Fig.~\ref{fig:tree_search} (left). We associate an empty sequence of grounded joint actions with node $D_0$, denoted as $D_0.\mathbf{S} = \emptyset$. We use the ``$.$" operator to denote the association relationship. This implies that at node $D_0$, we do not have any grounded joint actions. We then create edges originating from node $D_0$, with each edge storing a distinct initial task skeleton.}

\update{Throughout our search process, at each search iteration, we select an edge that has not been evaluated yet, and we evaluate it by trying to ground the task skeleton associated with it. As previously defined, the term \textit{grounding} refers to the process of finding feasible object placements and motion trajectories for a task skeleton to be executable. After each evaluation, we compute a reward based on the evaluation result. The reward will then be propagated back up the search tree, with each edge in the path from the root node to the selected edge having its value updated based on the reward. We use a Monte-Carlo Tree Search (MCTS) exploration strategy to balance exploration (exploring different unevaluated edges) and exploitation (biasing the search towards the branches that have received high rewards). }

\update{We use a reverse search algorithm inspired by~\cite{4209604} to ground task skeletons. We will describe our key component for task-skeleton grounding in detail in Sec.~\ref{sec:inst}. The insight behind the reverse search algorithm is to use the grounded future joint actions as the artificial constraints to guide the grounding for the current actions. Therefore, throughout our search process, we save the grounding results and use them as artificial constraints for subsequent grounding tasks. We use two examples, as shown in Fig.~\ref{fig:tree_search} (middle, right), to illustrate the idea.}

\update{In the first example (Fig.~\ref{fig:tree_search} (middle)), we select edge $E_2$ for evaluation. We create a new node, denoted as $D_2$, to serve as the head node of edge $E_2$. The tail node of edge $E_2$ is the root node $D_0$ whose associated sequence of grounded joint actions is empty. This means that we can attempt to ground the task skeleton associated with $E_2$, denoted as $E_2.\bar{\mathbf{S}}$, without any artificial constraints. Ideally, if we manage to ground task skeleton $E_2.\bar{\mathbf{S}}$ successfully, we would get an executable task-and-motion plan to perform the task. However, in many situations, we can only ground the task skeleton partially. This implies that there are conflicts that emerge during task-skeleton grounding. For example, there would not be enough space to place objects unless we relocate some objects that were not planned to be moved initially. Such situations can arise as we cannot account for all geometric specifics during task-skeleton generation. In such situations, we generate new task-skeletons to address the emerged conflicts, and we expand the tree by creating new edges, with each edge storing a distinct new task skeleton. In our first example, we create new edges originating from node $D_2$. Moreover, we store the sequence of joint actions that have been grounded to this point in node $D_2$, denoted as $D_2.\mathbf{S}$. It should be noted that $D_2.\mathbf{S}$ contains $D_0.\mathbf{S}$ and the grounded part of $E_2.\bar{\mathbf{S}}$.}

\update{In the second example (Fig.~\ref{fig:tree_search} (right)), we select edge $E_{2.1}$ for evaluation. The grounding of the task skeleton associated with edge $E_{2.1}$, denoted as $E_{2.1}.\bar{\mathbf{S}}$, should consider $D_2.\mathbf{S}$ as artificial constraints which is the sequence of joint actions that have been grounded to this point. If we successfully ground $E_{2.1}.\bar{\mathbf{S}}$, we can get an executable task-and-motion plan by concatenating the grounded task-skeleton with $D_2.\mathbf{S}$.}

\update{At each search iteration, we have four phases: \textit{selection}, \textit{expansion}, \textit{evaluation} and \textit{backpropagation}.}

\noindent\textbf{Notation.} We use $\lvert \mathbf{S}\rvert$ and $\lvert \bar{\mathbf{S}}\rvert$ to denote the number of objects intended to be moved in sequences of grounded joint actions $\mathbf{S}$ and task skeletons $\bar{\mathbf{S}}$, respectively.

\noindent\textbf{\textit{Selection} phase.} In the \textit{selection} phase, we start at the root node and recursively select the edge with the highest Upper Confidence Bound (UCB) value until we reach an edge $E_i$ with a task skeleton that has not been grounded yet. We denote the tail node of edge $E_i$ as $D_j$. We follow the UCB value formula used in~\cite{silver2017mastering}. The UCB value of the pair of node $D_j$ and edge $E_i$ is: $Q(D_j, E_i) = \frac{E_i.value}{E_i.visits + 1} + c \times E_i.prior \times \frac{\sqrt{D_{j}.visits}}{E_i.visits + 1}$, where $E_i.value$ is the cumulative reward edge $E_i$ has received so far, $D_j.visits$ and $E_i.visits$ are the number of times $D_j$ and $E_i$ have been selected, $c$ is a constant to balance exploration and exploitation, and $E_i.prior$ is used to bias the search with domain knowledge~\cite{silver2017mastering}. In our implementation, we set $E_{i}.prior$ to $\frac{1}{\lvert E_i.\bar{\mathbf{S}}\rvert}$ to prioritize grounding task skeletons with fewer objects to move. The value $E_i.value$ of an edge is initialized to $0$.

Assume that we select edge $E_i$ from node $D_j$ in the \textit{selection} phase. 

\noindent\textbf{\textit{Expansion} phase.} In the \textit{expansion} phase, we create a new node \update{$D_{i}$} as the head node of edge $E_i$.

\noindent\textbf{\textit{Evaluation} phase.} In the \textit{evaluation} phase, we use the task-skeleton grounding component (Sec.~\ref{sec:inst}) to ground task skeleton $E_i.\bar{\mathbf{S}}$ associated with $E_i$ to compute reward $r$ for selecting edge $E_i$. \update{Note that node $D_j$ is the tail node of edge $E_i$ and the grounded sequence of joint actions stored in node $D_j$ is denoted as $D_j.\mathbf{S}$.} There are three possible outcomes: (\romannum{1}) If we fail at grounding, we set $r$ to $0$. (\romannum{2}) If we obtain a sequence of grounded joint actions $\mathbf{S}^*$, then we found a valid task-and-motion plan. In this case, we set $r$ to $1 + \alpha \frac{1}{\lvert \mathbf{S}^*\rvert}$, where $\alpha$ is a constant hyperparameter used to balance the two terms of the reward that is set to $1$ in our experiments (Sec.~\ref{sec:exp}). The first term of the reward incentivizes the search algorithm to select edges where more actions have been grounded, and the second term incentivizes the search algorithm to select edges that move fewer objects. (\romannum{3}) \update{In the third case, task skeleton $E_i.\bar{\mathbf{S}}$ cannot be fully grounded without relocating some objects that are not planned to be moved in $E_i.\bar{\mathbf{S}}$. In this case, we obtain a sequence of grounded joint actions $\mathbf{S}'$ and a set of objects $\mathbf{M}^*$ from the grounding process. Here, $\mathbf{S}'$ consists of $D_j.\mathbf{S}$ and the grounded part of task skeleton $E_i.\bar{\mathbf{S}}$. We use $\mathbf{M}^*$ to represent the set of objects for which we need to find a sequence of grounded joint actions, denoted as $\mathbf{S}_{\mathbf{M}^*}$, to relocate so that we can construct a final task-and-motion plan for the problem by concatenating $\mathbf{S}_{\mathbf{M}^*}$ with $\mathbf{S}'$. We then} call the task-skeleton generating component (Sec.~\ref{sec:approx}) to move $\mathbf{M}^*$. If we cannot find any task skeleton to move $\mathbf{M}^*$, then we set $r$ to $0$. However, if we find a set of task skeletons $\{\bar{\mathbf{S}}\}$, then we set $r$ to $\frac{\mathbf{S}'.length}{\mathbf{S}'.length + \bar{\mathbf{S}}^*.length} + \alpha \frac{1}{\lvert \mathbf{S}'\rvert + \lvert \bar{\mathbf{S}}^* \rvert}$, where $\bar{\mathbf{S}}^*$ is the task skeleton with the minimum number of time steps among all task skeletons $\{\bar{\mathbf{S}}\}$ and $\mathbf{S}'.length$ and $\bar{\mathbf{S}}^*.length$ represent the number of time steps of $\mathbf{S}'$ and $\bar{\mathbf{S}}^*$, respectively. 

We would like to point out that the reward in the second possible outcome represents a special case of the reward in the third possible outcome. Both rewards use their first terms to incentivize the search algorithm to select edges where more actions have been grounded, and their second terms to incentivize the search algorithm to select edges that move fewer objects.

We use node \update{$D_{i}$} to store the returned grounded joint actions $\mathbf{S}'$ as \update{$D_{i}.\mathbf{S}$}. In the third scenario, if we find new task skeletons, then we create new edges to store them for node \update{$D_{i}$}. If no new edge is created, then we mark node \update{$D_{i}$} as a terminal node.

\noindent\textbf{\textit{Backpropagation} phase.} In the \textit{backpropagation} phase, we update the cumulative reward of the selected edges $\{E^{sel}\}$ with the computed reward $r$ according to $E^{sel}.value = {E^{sel}.value + r}$. We also increment the number of visits of the selected edges and nodes by $1$.

In our implementation, we track the grounding failures for different task skeletons similarly to~\cite{ren2021extended}, so that we can skip over those branches where grounding their task skeletons is known to be infeasible.

\subsubsection{Key Component 1: Generating Promising Task Skeletons}\label{sec:approx}
One key component in the second phase of our framework is to generate promising task skeletons $\{\bar{\mathbf{S}}\}$ for moving a set of objects $\mathbf{M}^*$ given a sequence of already grounded joint actions $\mathbf{S}'$. \update{As previously defined, the term \textit{task-skeleton} refers to a sequence of actions without the placement and trajectory information. This key component will be used in two situations. } It is \update{firstly} called at the initialization stage of the search process \update{(Fig.~\ref{fig:tree_search} (left)). In this situation, we set $\mathbf{S}'$ as empty and set $\mathbf{M}^*$ as the set of objects named in the goal specification of the problem instance. We will use the generated task skeletons to initialize the search tree as shown in Fig.~\ref{fig:tree_search} (left). The second scenario where this component is called is when we can only ground part of a task skeleton in the \textit{evaluation} phase during the search process. Fig.\ref{fig:tree_search} (middle) depicts one example search iteration where this situation happens. In this example search iteration, we set $\mathbf{S}'$ as $\mathbf{S}_2$ and set $\mathbf{M}^*$ as $\mathbf{M}_2^*$. We use this key component to generate task skeletons to relocate $\mathbf{M}_2^*$. We take $\mathbf{S}'$ as input because we should exclude objects from our task-skeleton generation that are already planned to be moved in $\mathbf{S}'$.} The task-skeleton generation algorithm is designed to utilize the computed collaborative manipulation information from the first phase (Sec.~\ref{sec:rep}) to eliminate task skeletons that include infeasible actions and to prioritize motion planning for effective task plans that have fewer time steps and fewer objects to be moved.

\noindent\textbf{Notation.} Assume that we want to generate task skeletons to move objects $\mathbf{M}^*$ given a sequence of grounded joint actions $\mathbf{S}'$. The set of objects included in $\mathbf{S}'$ cannot be moved again because of the \textit{monotone} assumption. For simplicity of presentation, we slightly abuse $\mathbf{M}$ to denote the movable objects not included in $\mathbf{S}'$.

\begin{figure*}[!t]
\centering
\includegraphics[width=0.7\linewidth]{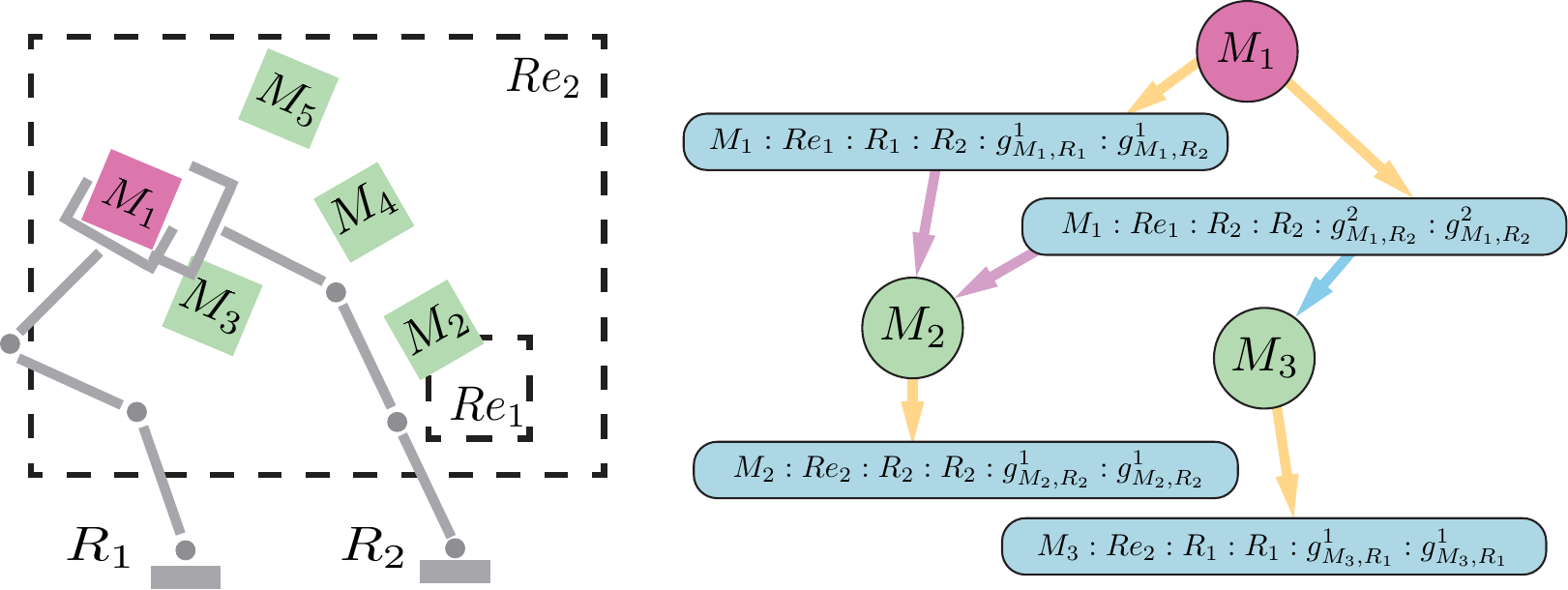}
\caption{(Left) An example scenario where we want to generate task skeletons to move object $\mathbf{M}_1$ given an empty sequence of grounded joint actions. (Right) The corresponding \textit{collaborative manipulation task graph} for moving object $M_1$. The rounded rectangular nodes are \textit{action nodes}. The circular nodes are \textit{object nodes}. The red circular nodes represent objects that are specified to be moved. The yellow arrows represent \textit{action edges}. The purple arrows represent \textit{block-place edges}, and the blue arrow represents a \textit{block-pick edge}.}
\label{fig:mg}
\end{figure*}

\noindent\textbf{Building the collaborative manipulation task graph.} To reason about the collaborative manipulation capabilities of the individual robots, we encode the computed information as a graph. We build a \textit{collaborative manipulation task graph} (CMTG) to capture the precedence of the manipulations of different objects, i.e., we can only move an object after we move the obstacles that block the pick-and-place action we are going to execute, based on the computed information from the first phase (Sec.~\ref{sec:rep}). Since we only compute occlusion information for placing objects named in the goal specification, the precedence encoded in the CMTG lack occlusion information for relocating objects that are not named in the goal specification. Instead, we assume that we will always find the feasible places to relocate these objects. We determine the exact object placements during task-skeleton grounding (Sec.~\ref{sec:inst}).

A CMTG (Fig.~\ref{fig:mg}) has two types of nodes: An \textit{object node} represents an object $M \in \mathbf{M}$; and an \textit{action node} represents a partially grounded pick-and-place action $\bar{a}$, i.e. a pick-and-place action without placement information.  A CMTG has three types of edges: An \textit{action edge} is an edge from an object node to an action node. It represents moving the object represented by the object node with the action represented by the action node. A \textit{block-pick edge} is an edge from an action node to an object node. It represents that the object represented by the object node obstructs the pick action of the action represented by the action node. A \textit{block-place edge} is an edge from an action node to an object node. It represents that the object represented by the object node obstructs the place action of the action represented by the action node. All \textit{block-place edges} are connected to the action nodes that move the objects named in the goal specification. A CMTG has a set of object nodes that represents the input objects $\mathbf{M}^*$ that must be moved.

Given the computed collaborative manipulation information and a set of objects $\mathbf{M}^{*}$ to move, we incrementally construct a CMTG by iteratively adding object $M \in \mathbf{M}^*$ to the CMTG with Alg.~\ref{alg:add_movable}. Given the CMTG $\mathbf{C}$ built so far and an object $M$ to add, we first add an object node representing $M$ to $\mathbf{C}$ (Alg.~\ref{alg:add_movable}, line~\ref{line:add_object_node}). Then, for each pair of a robot $R \in \mathbf{R}$ and its grasp $g_{M, R} \in \mathbf{Gr}_{M, R}$, we find all partially grounded pick-and-place actions $\mathbf{\bar{a}}$ that move object $M$ to its target region $Re_M$ with $R$ as the pick robot (Alg.~\ref{alg:add_movable}, line~\ref{line:get_goal_region}-\ref{line:get_actions}). For each partially grounded pick-and-place action $\bar{a}$, we find all movable objects that block the pick action of $\bar{a}$ and add the corresponding block-pick edges (Alg.~\ref{alg:add_movable}, line~\ref{line:pick_block}-\ref{line:pick_block_end}). If $M$ is named in goal specification $\mathcal{G}$, then we also find all movable objects that block the place action of $\bar{a}$ and add the corresponding block-place edges (Alg.~\ref{alg:add_movable}, line~\ref{line:place_block}-\ref{line:place_block_end}). We recursively add the blocking objects in a similar way (Alg.~\ref{alg:add_movable}, lines~\ref{line:explore_pick} and~\ref{line:explore_place}).

\setlength{\textfloatsep}{2pt}
\begin{algorithm*}[!t]
\caption{\textsc{AddObject}($M, \mathbf{C}$)}\label{alg:add_movable}
\begin{algorithmic}[1]
\State{\textbf{input}: an object $M$; the collaborative manipulation task graph built so far, denoted as $\mathbf{C}$.}
\If {$M \in \mathbf{C}.object\_nodes$}
\State{\Return}
\EndIf
\State {$\mathbf{C}.object\_nodes.add(M)$} \label{line:add_object_node}
\If {$M$ is named in goal specification $\mathcal{G}$}\label{line:get_goal_region}
\State {$Re_M$ = \textsc{GetGoalRegion}($M$)}
\Else 
\State {$Re_M$ = \textsc{GetCurrentRegion}($M$)}
\EndIf
\For {$R^{pick} \in \mathbf{R}$}
    \For {$g_{M, R^{pick}} \in \mathbf{Gr}_{M, R^{pick}}$}
        \State{$\mathbf{\bar{a}} = \{\}$}
        \If {\textsc{ReachablePick}$(M, g_{M, R^{pick}}, R^{pick})$}
             \If {\textsc{ReachablePlace}$(M, Re_M, g_{M, R^{pick}}, R^{pick})$}
                \State{$\mathbf{\bar{a}}.add((M, Re_M, R^{pick},R^{pick}, g_{M, R^{pick}},g_{M, R^{pick}}))$}
             \EndIf
            \If {$M$ is named in goal specification $\mathcal{G}$}
                \For {$R^{place} \in \mathbf{R} \setminus \{R^{pick}\}$}
                    \For {$g_{M, R^{place}} \in \mathbf{Gr}_{M, R^{place}}$}
                        \If {\textsc{EnableGoalHandover}$(M, g_{M, R^{pick}}, g_{M, R^{place}}, R^{pick}, R^{place})$ and \\ \hspace{11em}\textsc{ReachablePlace}$(M, Re_M, g_{M, R^{place}}, R^{place})$}
                            \State {$\mathbf{\bar{a}}$.add($(M, Re_M, R^{pick},R^{place}, g_{M, R^{pick}},g_{M, R^{place}})$)}
                        \EndIf
                    \EndFor
                \EndFor
            \EndIf\label{line:get_actions}
            \For {$\bar{a} \in \mathbf{\bar{a}}$}
                \State{$R^{pick}_{\bar{a}}$ is the robot to pick $M$ in $\bar{a}$}
                \State{$g^{pick}_{\bar{a}}$ is the grasp used by $R^{pick}_{\bar{a}}$ in $\bar{a}$}
                \State{$R^{place}_{\bar{a}}$ is the robot to place $M$ in $\bar{a}$}
                \State{$g^{place}_{\bar{a}}$ is the grasp used by $R^{place}_{\bar{a}}$ in $\bar{a}$}
                
                \State{$\mathbf{C}.action\_nodes.add(\bar{a})$}
                \State{$\mathbf{C}.action\_edges.add(M, \bar{a})$}
                
                \For {$M_j \in \mathbf{M}$}\label{line:pick_block}
                    \If {\textsc{OccludesPick}$(M_j, M, g^{pick}_{\bar{a}}, R^{pick}_{\bar{a}})$}
                        \State {\textsc{AddObject}($M_j, \mathbf{C}$)}\label{line:explore_pick}
                        \State {$\mathbf{C}.block\_pick\_edges.add(\bar{a}, M_j)$}
                    \EndIf
                \EndFor\label{line:pick_block_end}

                \If{$M$ is named in goal specification $\mathcal{G}$}\label{line:place_block}
                \For {$M_j \in \mathbf{M}$}
                    \If{\textsc{OccludesGoalPlace}$(M_j, M, Re_M, g^{place}_{\bar{a}}, R^{place}_{\bar{a}})$}
                        \State {\textsc{AddObject}($M_j, \mathbf{C}$)}\label{line:explore_place}
                        \State {$\mathbf{C}.block\_place\_edges.add(\bar{a}, M_j)$}
                    \EndIf 
                \EndFor  
                \EndIf\label{line:place_block_end}
            \EndFor 
        \EndIf 
    \EndFor 
\EndFor

\end{algorithmic}
\end{algorithm*}

\noindent\textbf{Mixed-integer linear program formulation and solving.} Given a CMTG $\mathbf{C}$, we find a set of task skeletons that specify which robot will move which object at each time step. We assume that each object will be moved at most once, i.e., we assume that the problem instances are \textit{monotone}. Given a time step limit $T$, we cast the problem of finding a task skeleton that has a minimum number of objects to be moved as a mixed-integer linear program (MIP). We encode the precedence of manipulating different objects as formal constraints in the MIP such that we can generate task skeletons that are promising to be successfully grounded. We incrementally increase the time step limit $T$. In our implementation, the maximum time step limit is a hyperparameter.

For simplicity of presentation, we slightly abuse $\mathbf{M}$ again to denote the objects in $\mathbf{C}$. We use $\mathbf{M}^* \subseteq \mathbf{M}$ to denote the objects that are intended to be moved. We slightly abuse $\mathbf{\bar{a}}$ to denote the set of partially grounded pick-and-place actions in $\mathbf{C}$. We use $E_{\mathbf{\bar{a}}} = \{(M, \bar{a})\}$ to denote the set of action edges in $\mathbf{C}$. We use $E^{pick}_B = \{(\bar{a}, M)\}$ to denote the set of block-pick edges and $E^{place}_B = \{(\bar{a}, M)\}$ to denote the set of block-place edges in $\mathbf{C}$, $E_B = E^{pick}_B \cup E^{place}_B$,  where $M \in \mathbf{M}$ and $\bar{a} \in \mathbf{\bar{a}}$. We define the binary variables $X^t_{M, \bar{a}}$ and $X^t_{\bar{a}, M}$, where $t \in [1, \dots, T], (M, \bar{a}) \in E_{\mathbf{\bar{a}}}$ and $(\bar{a}, M) \in E_B$. $X^t_{M, \bar{a}} = 1$ implies that action $\bar{a}$ is executed at time step $t'\text{ s.t. }t' \geq t$. $X^t_{\bar{a}, M} = 1$ implies that object $M$ can be considered for being moved at time step $t$ since it blocks action $\bar{a}$ which is executed at or after time step $t$.

Our MIP model is shown in the following. The implications in constraint $(11)$ and constraint $(12)$ are compiled to linear constraints using the big-M method~\cite{griva2009linear}:
\setlength{\abovedisplayskip}{3pt}%
\setlength{\belowdisplayskip}{3pt}%
\setlength{\abovedisplayshortskip}{0pt}%
\setlength{\belowdisplayshortskip}{0pt}%
\setlength{\jot}{2pt}%
{
\scriptsize
\begin{align}
&\text{\normalsize{minimize}} \sum\nolimits_{(M, \bar{a}) \in E_{\mathbf{\bar{a}}}} X^1_{M, \bar{a}} \nonumber \\
&X^{t}_{M, \bar{a}} \geq X^{t + 1}_{M, \bar{a}}, \forall (M, \bar{a}) \in E_{\mathbf{\bar{a}}}, t \in [1, T - 1] \\
&X^t_{M, \bar{a}} = X^t_{\bar{a}, M'}, \forall (M, \bar{a}) \in E_{\mathbf{\bar{a}}}, (\bar{a}, M') \in E_B, \nonumber \\ & \hspace{16em} t \in [1, T] \\
&X^t_{M, \bar{a}'} \leq \sum\nolimits_{(\bar{a}, M) \in E_{B}} X^t_{\bar{a}, M}, \forall M \in \mathbf{M} \setminus \mathbf{M}^*, \nonumber \\ & \hspace{10em} (M, \bar{a}') \in E_{\mathbf{\bar{a}}}, t \in [1, T] \\
&\sum\nolimits_{(M, \bar{a}) \in E_{\mathbf{\bar{a}}} \text{ s.t. } R \text{ in } \bar{a}} X^T_{M, \bar{a}} \leq 1, \forall R \in \mathbf{R} \\
&\sum\nolimits_{(M, \bar{a}) \in E_{\mathbf{\bar{a}}}} X^T_{M, \bar{a}} \geq 1 \\
&\sum\nolimits_{(M, \bar{a}) \in E_{\mathbf{\bar{a}}} \text{ s.t. } R \text{ in } \bar{a}} X^t_{M, \bar{a}} \leq 1 + \nonumber \\ & \hspace{7em} \sum\nolimits_{(M, \bar{a}) \in E_{\mathbf{\bar{a}}} \text{ s.t. } R \text{ in } \bar{a}} X^{t + 1}_{M, \bar{a}}, \nonumber \\ & \hspace{10em} \forall R \in \mathbf{R}, t \in [1, T - 1] \\
&\sum\nolimits_{(M, \bar{a}) \in E_{\mathbf{\bar{a}}}} X^t_{M, \bar{a}} \geq 1 + \sum\nolimits_{(M, \bar{a}) \in E_{\mathbf{\bar{a}}}} X^{t + 1}_{M, \bar{a}}, \nonumber \\ & \hspace{14em} t \in [1, T - 1] \\
&\sum\nolimits_{(M, \bar{a}) \in E_{\mathbf{\bar{a}}}} X^1_{M, \bar{a}} = 1, \forall M \in \mathbf{M}^* \\
&\sum\nolimits_{(M, \bar{a}') \in E_{\mathbf{\bar{a}}}} X^1_{M, \bar{a}'} \geq X^1_{\bar{a}, M},  \forall (\bar{a}, M) \in E_B\\
&\sum\nolimits_{(M, \bar{a}) \in E_{\mathbf{\bar{a}}}} X^1_{M, \bar{a}} \leq 1, \forall M \in \mathbf{M} \\
& X^1_{\bar{a}, M} = 1 \implies \sum\nolimits_{t \in [1, \dots, T]} X^t_{\bar{a}, M} \geq \nonumber \\ & \hspace{3em} (\sum\nolimits_{(M, \bar{a}') \in E_{\mathbf{\bar{a}}}} \sum\nolimits_{t \in [1, \dots, T]} X^t_{M, \bar{a}'}) + 1, \nonumber \\ & \hspace{13em} \forall (\bar{a}, M) \in E^{pick}_B \\
& X^1_{\bar{a}, M} = 1 \implies \sum\nolimits_{t \in [1, \dots, T]} X^t_{\bar{a}, M} \geq \nonumber \\ & \hspace{5em} (\sum\nolimits_{(M, \bar{a}') \in E_{\mathbf{\bar{a}}}} \sum\nolimits_{t \in [1, \dots, T]} X^t_{M, \bar{a}'}), \nonumber \\ & \hspace{13em} \forall (\bar{a}, M) \in E^{place}_B
\end{align}
\normalsize
}

Constraint $(1)$ enforces that $X^t_{M, \bar{a}}$ indicates whether we have selected $\bar{a}$ at or after time step $t$. Constraint $(2)$ enforces that, if an action is selected, then the objects that obstruct it are also moved. Constraint $(3)$ enforces that, besides the objects in $\mathbf{M}^*$, we only move objects that obstruct the actions we have selected. Constraints $(4-7)$ enforce that, at each time step, we select at least one action, while each robot executes at most one action. Constraint $(8)$ enforces that the objects in $\mathbf{M}^*$ are moved. Constraint $(9)$ enforces that all obstacles for the selected actions are moved, while constraint $(10)$ enforces that each object is moved only once. Constraint $(11)$ enforces that each object is moved after the obstacles for its pick action have been moved. Constraint $(12)$ enforces that each object is moved after the obstacles for its place action have been moved. The objective function represents the number of moved objects.

From a MIP solution, we construct a task skeleton which is grounded later. Moreover, we want to construct multiple task skeletons since some task skeletons may be impossible to ground. Every time we obtain a solution, we add a constraint to the MIP model to enforce that we find a different solution from the existing ones until we collect enough task skeletons~\cite{10.1007/978-3-540-72792-7_22}. In our implementation, the maximum number of task skeletons is a hyperparameter that varies for different problem instances.

\subsubsection{Key Component 2: Task-Skeleton Grounding}\label{sec:inst}
\setlength{\textfloatsep}{2pt}
\begin{algorithm*}[!t]
\caption{\textsc{\update{Task-Skeleton Grounding}}\update{($\bar{\mathbf{S}}, \mathbf{S}_{fut}, \mathbf{M}_{fut}, \mathbf{V}_{fut}, \mathbf{M}_{out}$)}}\label{alg:grounding}
\begin{algorithmic}[1]
\State{\textbf{input}: a task skeleton $\bar{\mathbf{S}}$; a sequence of grounded joint actions $\mathbf{S}_{fut}$; the set of objects that are planned \hspace*{2.9em} to be moved in $\mathbf{S}_{fut}$, denoted as $\mathbf{M}_{fut}$; the volume of work space that is occupied by $\mathbf{S}_{fut}$, \hspace*{2.9em} denoted as $V_{fut}$; the set of movable objects that are not planned to be moved in $\mathbf{S}_{fut}$ and $\bar{\mathbf{S}}$, \hspace*{2.9em} denoted as $\mathbf{M}_{out}$.}
\State{\textbf{result:} three possible returns: (\romannum{1}) a sequence of grounded joint actions $\mathbf{S}^*$, indicating that we \hspace*{3.1em} successfully find an executable task-and-motion plan; (\romannum{2}) a sequence of grounded joint \hspace*{3.3em} actions $\mathbf{S}'$ and a set of objects $\mathbf{M}^*$, indicating that we can only partially ground task skeleton \hspace*{3.3em} $\bar{\mathbf{S}}$ and we have to relocate objects $\mathbf{M}^*$; (\romannum{3}) a failure flag.}
\State {\textbf{notation:} We denote the sequence concatenating operation as $\oplus$.}
\State {$\mathcal{G}$ = goal specification of the MR-GTAMP problem instance}
\State {$\mathbf{M}$ = the set of movable objects of the MR-GTAMP problem instance}
\For {$t \in [T, \dots, 1]$}
    \State {$\bar{\mathbf{S}}[t]$ = \textsc{PartiallyGroundedJointActionAt}($\bar{\mathbf{S}}, t$)}
    \State {$\mathbf{M}^t, \mathbf{R}^t$ = \textsc{ObjectsAndRobotsToMove}($\bar{\mathbf{S}}[t]$)}
    \State {$\mathbf{P}$ = \textsc{FindPlacements}($\mathbf{M}^t$, $\mathbf{M}_{out} \cup \mathbf{M}_{fut} \cup \mathbf{F} \cup V_{fut}$, $ \bar{\mathbf{S}}[t]$)} \label{line:placement_w_out}
    \If {$\mathbf{P}$ is None} \label{line:placement_wo_out_start}
        \State {$\mathbf{P}$ = \textsc{FindPlacements}($\mathbf{M}^t$, $\mathbf{M}_{fut} \cup \mathbf{F} \cup V_{fut}$, $ \bar{\mathbf{S}}[t]$)}
        \If {$\mathbf{P}$ is None}
            \State{\Return {failure flag}}
        \EndIf
        \State {$\Xi$ = \textsc{FindTrajectories}($\mathbf{M}^t$, $\mathbf{R}^t$, $\mathbf{P}$, $\mathbf{M}_{fut} \cup \mathbf{F}$, $ \bar{\mathbf{S}}[t]$)}
        \If {$\Xi$ is None}
            \State{\Return {failure flag}}
        \EndIf
        \State {$\mathbf{S}_{fut}$ = \textsc{CreateGroundedJointAction}$(\bar{\mathbf{S}}[t], \Xi, \mathbf{P}) \oplus \mathbf{S}_{fut}$}
        \State {$\mathbf{M}^*$ = \textsc{HaveNotBeenMoved}($\mathcal{G}$, $\mathbf{S}_{fut}$) $\cup$ \textsc{MovablesOcclude}($\mathbf{M}$, $\mathbf{S}_{fut}$)} \label{line:new_m_1}
        \State {$\mathbf{S}'$ = $\mathbf{S}_{fut}$}
        \State{\Return {$\mathbf{S}', \mathbf{M}^*$}}
    \EndIf \label{line:placement_wo_out_end}
    \State {$\Xi$ = \textsc{FindTrajectories}($\mathbf{M}^t$, $\mathbf{R}^t$, $\mathbf{P}$, $\mathbf{M}_{out} \cup \mathbf{M}_{fut} \cup \mathbf{F}$, $ \bar{\mathbf{S}}[t]$)} \label{line:traj_w_out}
    \If {$\Xi$ is None} \label{line:traj_wo_out_start}
        \State {$\Xi$ = \textsc{FindTrajectories}($\mathbf{M}^t$, $\mathbf{R}^t$, $\mathbf{P}$, $\mathbf{M}_{fut} \cup \mathbf{F}$, $ \bar{\mathbf{S}}[t]$)}
        \If {$\Xi$ is None}
            \State{\Return {failure flag}}
        \EndIf
        \State {$\mathbf{S}_{fut}$ = \textsc{CreateGroundedJointAction}$(\bar{\mathbf{S}}, t, \Xi, \mathbf{P}) \oplus \mathbf{S}_{fut}$}
        \State {$\mathbf{M}^*$ = \textsc{HaveNotBeenMoved}($\mathcal{G}$, $\mathbf{S}_{fut}$) $\cup$ \textsc{MovablesOcclude}($\mathbf{M}$, $\mathbf{S}_{fut}$)} \label{line:new_m_2}
        \State {$\mathbf{S}'$ = $\mathbf{S}_{fut}$}
        \State{\Return {$\mathbf{S}', \mathbf{M}^*$}} 
    \EndIf \label{line:traj_wo_out_end}
    \State {$\mathbf{M}_{fut}$ = $\mathbf{M}_{fut} \cup \mathbf{M}^{t}$} \label{line:expand_start}
    \State {$\mathbf{S}_{fut}$ = \textsc{CreateGroundedJointAction}$(\bar{\mathbf{S}}, t, \Xi, \mathbf{P}) \oplus \mathbf{S}_{fut}$}
    \State {$V_{fut}$ = $V_{fut}$.append(\textsc{SweptVolume($\Xi, \mathbf{M}^{t}, \mathbf{R}^t$)})} \label{line:expand_end}
\EndFor
\State {$\mathbf{S}^* = \mathbf{S}_{fut}$}
\State{\Return {$\mathbf{S}^*$}}
\end{algorithmic}
\end{algorithm*}
The second key component in the search phase (Sec.~\ref{sec:tree_search}) is to ground the task skeletons, i.e., to find the object placements and motion trajectories for the partially grounded pick-and-place actions. We use a reverse search algorithm inspired by~\cite{4209604} since forward search for continuous parameters of long-horizon task skeletons without any guidance is very challenging~\cite{Kim2019}. The insight behind the reverse search strategy is to use the grounded future joint actions as the artificial constraints to guide the grounding for the present time step.

The input to this component is a task skeleton $\bar{\mathbf{S}}$ of $T$ time steps and a sequence $\mathbf{S}_{fut}$ of future grounded joint actions. \update{We use $\mathbf{S}_{fut}$ as artificial constraints to guide the grounding for the current actions, so that we can efficiently find geometrically feasible long-horizon plans~\cite{4209604}. Ideally, if we manage to ground task skeleton $\bar{\mathbf{S}}$ successfully, we will get a fully executable task-and-motion plan. However, in many situations, since we cannot account for all geometric specifics during task-skeleton generation, we can only ground the task skeleton partially. In such cases, we will get a set of objects, denoted as $\mathbf{M}^*$, for which we have to generate new task skeletons to relocate. We will then return the sequence of grounded joint actions together with objects $\mathbf{M}^*$. Furthermore, in certain situations, the grounding may totally fail. In such cases, we will simply return a failure flag.} 

We denote the volume of work space occupied by grounded joint actions $\mathbf{S}_{fut}$ as $V_{fut}$. We denote the set of movable objects that will be moved by grounded joint actions $\mathbf{S}_{fut}$ as $\mathbf{M}_{fut}$. We denote the set of movable objects that will not be moved by task skeleton $\bar{\mathbf{S}}$ and grounded joint actions $\mathbf{S}_{fut}$ as $\mathbf{M}_{out}$. For time step $t \in [1, \dots, T]$, we denote the set of objects that are planned to be moved as $\mathbf{M}^t$ and the set of robots that are planned to move them as $\mathbf{R}^t$. Recall that we denote the goal specification and the set of movable objects as $\mathcal{G}$ and $\mathbf{M}$, respectively.

\update{The detailed grounding algorithm is as follows (Alg.~\ref{alg:grounding})}. The grounding starts at the last time step $T$. For time step $t$, we first sample placements for objects $\mathbf{M}^t$ that are collision-free with respect to objects $\mathbf{M}_{out} \cup \mathbf{M}_{fut}$ \update{at their initial poses}, fixed objects $\mathbf{F}$ and volume $V_{fut}$ \update{(Alg.~\ref{alg:grounding}, line~\ref{line:placement_w_out})}. The sampled placements should not collide with volume $V_{fut}$, because, otherwise, they will prevent the execution of future grounded joint actions that occupy $V_{fut}$. 

Given the placements, we plan pick trajectories, place trajectories and handover trajectories for objects $\mathbf{M}^t$ and robots $\mathbf{R}^t$ that are collision-free with respect to objects $\mathbf{F} \cup \mathbf{M}_{fut} \cup \mathbf{M}_{out}$ \update{at their initial poses} \update{(Alg.~\ref{alg:grounding}, line~\ref{line:traj_w_out})}. We note that, in addition to the fixed objects $\mathbf{F}$ and the objects $\mathbf{M}_{out}$, the planned trajectories should not collide with the objects $\mathbf{M}_{fut}$ that are moved in future grounded joint actions.

Since we may move multiple robots and objects concurrently, we do not allow collisions between the robots, collisions between the moved objects and collisions between a robot and a moved object that is not intended to be manipulated by that robot. 

If we succeed in grounding the joint action at time step $t$, then we expand volume $V_{fut}$ with the volume occupied by the newly planned robot and object trajectories, expand the set $\mathbf{M}_{fut}$ with the moved objects $\mathbf{M}^t$ and expand the grounded joint actions $\mathbf{S}_{fut}$ with the newly grounded joint action \update{(Alg.~\ref{alg:grounding}, line~\ref{line:expand_start}-\ref{line:expand_end})}. We then start to ground the joint action at time step $t - 1$. If we succeed in grounding the joint actions at every time step, we return an executable task-and-motion plan $\mathbf{S}^* = \mathbf{S}_{fut}$. However, if we fail at grounding the joint action at time step $t$, we relax the collision constraints by allowing the sampled placements and trajectories to collide with the objects $\mathbf{M}_{out}$ since we can generate new skeletons to move them later \update{(Alg.~\ref{alg:grounding}, line~\ref{line:placement_wo_out_start}-\ref{line:placement_wo_out_end} and line~\ref{line:traj_wo_out_start}-\ref{line:traj_wo_out_end})}. If we succeed after relaxing the constraints, then we terminate the grounding and return the sequence of the grounded joint actions $\mathbf{S}' = \mathbf{S}_{fut}$ and a set of objects $\mathbf{M}^*$. The set of objects $\mathbf{M}^*$ consists of the objects that are named in the goal specification $\mathcal{G}$ but have not yet been moved and the movable objects in the environment that occlude the grounded joint actions $\mathbf{S}'$ \update{(Alg.~\ref{alg:grounding}, line~\ref{line:new_m_1} and line~\ref{line:new_m_2})}. During the search process (Sec.~\ref{sec:tree_search}), the returned $\mathbf{S}'$ and $\mathbf{M}^*$ are then used as input to the first key component (Sec.~\ref{sec:approx}) to generate new task skeletons. If, after relaxing the collision constraints, we still cannot find feasible placements and paths, then we simply return failure.

\section{Experiments}\label{sec:exp}

We empirically evaluate our framework in two challenging domains and show that it can generate effective collaborative task-and-motion plans more efficiently than two baselines.

\subsection{Baselines}

We compare our framework with two state-of-the-art TAMP frameworks. We provide both baseline planners with information about the reachable regions of each robot.

Ap1 is a multi-robot extension of the RSC algorithm~\cite{4209604} by assuming that the robots form a single composite robot. The action space includes all possible combinations of the single-robot actions and collaboration actions. \update{Unlike our framework, which eliminates infeasible task plans using computed information about the manipulation capabilities of individual robots (Sec.~\ref{sec:rep}), thereby pruning the search space, Ap1 would require searching through a large space of all possible combinations of multi-agent actions. Moreover, the focus of Ap1 is on feasibility of the task-and-motion plans, rather than on the plan length and number of objects moved. In contrast, our framework uses the intermediate grounding results (Sec.~\ref{sec:tree_search}) to guide the  search towards more effective task-and-motion plans, considering the resulting plan length and the number of objects moved.}

Ap2 is a general MR-TAMP framework~\cite{9636119} that is efficient in searching for promising task plans based on the constraints incurred during motion planning. We implement the planner in a way such that geometric constraints can be utilized efficiently, e.g., the planner can identify that it needs to move the blocking objects away before it can manipulate the blocked objects. \update{Unlike our framework, which guides the search for feasible positions for object relocation using sampled future actions (Sec.~\ref{sec:inst}), Ap2 does not include guidance for finding feasible positions for object relocation, which can facilitate finding feasible plans in confined settings.}

\begin{table*}[ht]\centering
\Large
\caption{Comparison of the proposed method with two baseline methods in the two benchmark domains regarding the success rate, planning time, makespan and motion cost. The numbers in the names of the problem instances indicate the numbers of the goal objects and the movable objects besides the goal objects. In PA5, PA7 and PA10, each problem instance has $3$ goal objects and $2$ robots. We omit the planning time and solution quality results for Ap2 on PA10 \update{and BO8} because its success rate is significantly lower than those of the other two methods.}
\resizebox{1.\linewidth}{!}{
\begin{tabular}{ccccccccccccc}
    \hline
    \\ [-1em]
    Problem instance & \multicolumn{3}{c}{Success rate \%} & \multicolumn{3}{c}{Planning time (s)} & \multicolumn{3}{c}{Makespan} & \multicolumn{3}{c}{Motion cost} \\
    \\ [-1em]
    \cmidrule(lr){2-4} \cmidrule(lr){5-7} \cmidrule(lr){8-10} \cmidrule(lr){11-13}
    & Ap1 & Ap2 & Ours & Ap1 & Ap2 & Ours  & Ap1 & Ap2 & Ours & Ap1 & Ap2 & Ours \\
    \\ [-1em]
    \hline
    \\ [-1em]
    PA5 & \textbf{100.0} & 80.0 & \textbf{100.0} & 5.6 ($\pm$1.3) & 6.1 ($\pm$2.1) & \textbf{2.4 ($\pm$0.2)}  & 3.0 ($\pm$0.2) & 2.9 ($\pm$0.2) & \textbf{2.8 ($\pm$0.2)}  & 3.8 ($\pm$0.2) & \textbf{3.6 ($\pm$0.2)} & \textbf{3.6 ($\pm$0.2)} \\
    \\ [-1em]
    PA7 & 80.0 & 70.0 & \textbf{100.0} &  39.8 ($\pm$12.8) & 10.5 ($\pm$2.9) & \textbf{4.0 ($\pm$0.9)} & 3.7 ($\pm$0.3) & \textbf{3.0 ($\pm$0.3)} & 3.1 ($\pm$0.2) & 4.8 ($\pm$0.3) & 4.3 ($\pm$0.2) & \textbf{4.1 ($\pm$0.2)}\\
    \\ [-1em]
    PA10 & 55.0 & 40.0 & \textbf{90.0} & 129.2 ($\pm$58.2) & N/A & \textbf{19.6 ($\pm$6.1)}  & 4.6 ($\pm$0.6) & N/A & \textbf{4.2 ($\pm$0.3)} & 5.6 ($\pm$0.6) & N/A & \textbf{5.2 ($\pm$0.4)}\\
    \\ [-1em]
    BO8 & 85.0 & \update{35.0} & \textbf{100.0} & 246.5 ($\pm$54.2) & N/A & \textbf{182.2 ($\pm$48.3)}  & 4.8 ($\pm$0.2) & N/A & \textbf{3.4 ($\pm$0.3)} & 7.6 ($\pm$0.1) & N/A & \textbf{5.0 ($\pm$0.6)}\\

    \hline 
\end{tabular}
}
\label{tab:exp_time}
\end{table*}

\subsection{Benchmark Domains}

We evaluate the efficiency and effectiveness of our method and the two baselines in the \textbf{packaging} domain shown in Fig.~\ref{fig:best} (left) and the \textbf{box-moving} domain shown in Fig.~\ref{fig:best} (right).

\noindent\textbf{Packaging (PA):} In this domain, each problem instance includes $2$ to \update{$6$} robots, $3$ to $5$ goal objects, $2$ to $13$ movable objects besides the goal objects, $1$ start region and $3$ goal regions. As in~\cite{doi:10.1177/02783649211038280}, we omit motion planning and simply check for collisions at the picking and placing configurations computed by inverse kinematics solvers in this domain, because collisions in this domain mainly constrain the space of feasible picking and placing configurations. We use Kinova Gen2 lightweight robotic arms. For each benchmark problem instance, we conduct $20$ trials with a timeout of $1,200$ seconds. For all methods, we also count a trial as failed, if all possible task plans have been tried.

\noindent\textbf{Box-moving (BO):} \update{In this domain, we evaluate our framework for mobile manipulating tasks where the robots have to move target objects from one room to the other room (Fig.~\ref{fig:best} (right)). We use simulated PR2 robots.} In this domain, each problem instance includes $2$ robots, $2$ goal objects, $6$ movable objects besides the goal objects, $1$ start region and $1$ goal region. For simplification, we do not consider handover actions. For each benchmark problem instance, we conduct $20$ trials with a timeout of $1,200$ seconds. For both methods, we also count a trial as failed, if all possible task plans have been tried.

We use bidirectional rapidly-exploring random trees~\cite{lavalle2006planning} for motion planning and IKFast~\cite{diankov_thesis} for inverse kinematics solving. All methods share the same grasp sets, the same sets of single-robot actions and the same sets of collaboration actions. All experiments were run on an AMD Ryzen Threadripper PRO 3995WX Processor with a memory of 64GB.

\subsection{Results}
We refer to the number of time steps as \textit{makespan} and the number of moved objects as \textit{motion cost}.

\noindent\textbf{Planning time and success rate.} Table~\ref{tab:exp_time} shows that our method outperforms both baseline methods on all problem instances with different numbers of goal objects and movable objects with respect to both the planning times and success rates. Ap1 and our method achieved higher success rates on all problem instances than Ap2 because the reverse search strategy (Sec.~\ref{sec:inst}) utilized in Ap1 and our method can find feasible object placements much more efficiently than the forward search strategy used in Ap2. Moreover, Ap2 can generate task plans that include irrelevant objects while Ap1 and our method focus on manipulating the important objects, like blocking objects for necessary manipulation or goal objects. Our method achieved higher success rates with shorter planning times than Ap1 on the difficult problem instances PA7, PA10 and BO8 because our method first generates promising task skeletons (Sec.~\ref{sec:approx}) that use the information about the collaborative manipulation capabilities of the individual robots to prune the task plan search space, which can be extremely large when there are many objects and multiple robots~\cite{9636119}. The main cause of failure of our method is running out of task skeletons which can be addressed by incrementally adding more task skeletons during the search process.

\noindent\textbf{Solution quality.} Table~\ref{tab:exp_time} shows that our method can generate effective task-and-motion plans with respect to the motion cost and the makespan. Our method first generates task skeletons with short makespans by incrementally increasing time step limit and with low motion costs by incorporating the motion cost into the objective function of the MIP formulation (Sec.~\ref{sec:approx}). On the other hand, our MCTS exploration strategy motivates the planner to search for effective plans with small numbers of moved objects. It should be noted that, although Ap2 generated plans with shorter makespans for PA7, it has lower success rates and longer planning times than our method. Also, Ap1 generated plans that move significantly more objects for PA7, PA10 and BO8 than our method because it uses a depth-first search strategy for finding feasible plans~\cite{4209604}. 

\begin{table}[ht]\centering
 \Large
\caption{The results of the proposed method in domain PA regarding the success rate, planning time, makespan and motion cost. The numbers in the names of the problem instances indicate the numbers of the robots.}
\resizebox{1.\columnwidth}{!}{\begin{tabular}{ccccc}

    \hline
    \\ [-1em]
    Problem instance & \multicolumn{1}{c}{Success rate \%} & \multicolumn{1}{c}{Planning time (s)} & \multicolumn{1}{c}{Makespan} & \multicolumn{1}{c}{Motion cost} \\
    \cmidrule(lr){1-1}
    \cmidrule(lr){2-2} \cmidrule(lr){3-3} \cmidrule(lr){4-4} \cmidrule(lr){5-5} 
    \\ [-1em]
    2 robots & 60.0 & 148.4 ($\pm$36.8) & 6.1 ($\pm$0.4) & 8.9 ($\pm$0.4)\\
    \\ [-1em]
    3 robots & 80.0 & 99.0 ($\pm$48.6) & 4.9 ($\pm$0.3) & 8.2 ($\pm$0.5) \\
    \\ [-1em]
    4 robots & 85.0 & 109.1 ($\pm$33.6) & 4.7 ($\pm$0.3) & 8.2 ($\pm$0.4) \\
    \\[-1em]
    \update{5 robots} & \update{75.0} & \update{207.0} (\update{$\pm$48.7}) & \update{4.1 ($\pm$ 0.2)} & \update{8.0 ($\pm$ 0.3)}  \\
    \\ [-1em]
    \update{6 robots} & \update{70.0} & \update{362.7 ($\pm$ 64.6)} & \update{3.4 ($\pm$ 0.2)} & \update{7.7 ($\pm$ 0.4)} \\
    \\ [-1em]
    \hline 
\end{tabular}}
\label{tab:scale}
\end{table}

\noindent\textbf{\update{Scalability evaluation.}} \update{We evaluated the scalability of our method in the PA domain with $18$ movable objects, including $5$ goal objects and $2$ to \update{$6$} robots. Table~\ref{tab:scale} shows that our method can solve these large problem instances. For problem instances with $3$ and more robots, it achieved higher success rates compared to the problem instances with $2$ robots. Moreover, our method can achieve shorter makespans and lower motion costs when more robots are involved. These results show that our method can effectively utilize multiple robots to address challenging planning problem instances and generate intelligent collaboration strategies for multiple robots.}

\update{However, in our experiment, the success rates for problem instances with $5$ and $6$ robots are lower than the success rates for problem instances with $3$ and $4$ robots. The required planning time also increases when more robots are added, starting from the problem instances with $3$ robots. This is because adding more robots into the system will lead to more cluttered environments and more difficult collision avoidance between robots. In future work we will explore potentially mitigating this issue by carefully designing the layout of robots~\cite{tay1996optimising}.}

\section{Application Study: Roof Bolting}

\begin{figure}[ht]
\centering

\captionsetup{justification=centering}

\includegraphics[width=0.95\linewidth]{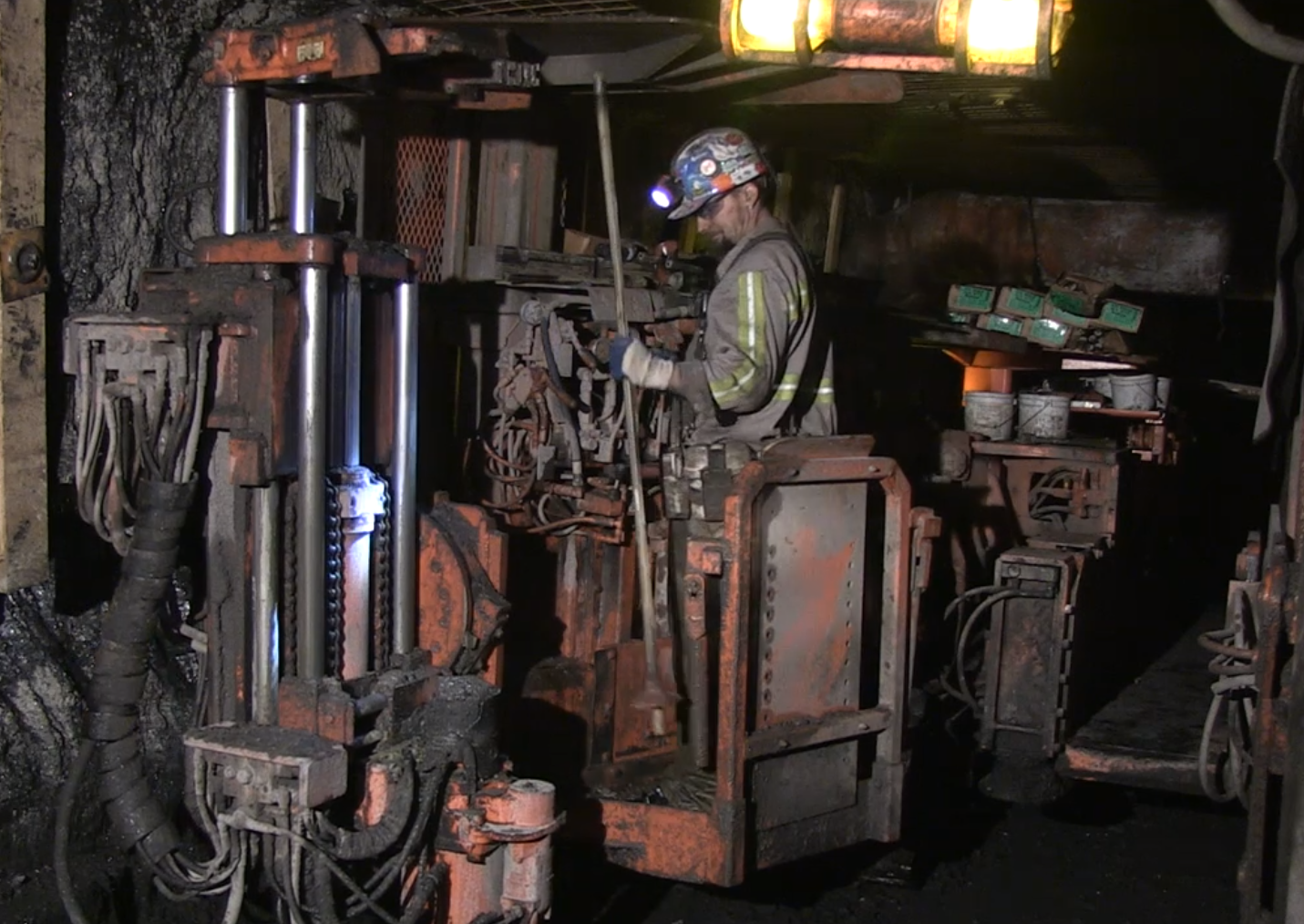}
\caption{A human operator is installing a bolt into the roof bolter.\protect\footnotemark}
\label{fig:human_mine}
\end{figure}

\footnotetext{\url{https://bit.ly/3tfYOMY}}

Roof bolting is an essential operation within the underground mining cycle, as it aims to provide support to the exposed roof and ribs of the new excavation~\cite{peng1984roof,mark2002introduction} (Fig.~\ref{fig:human_mine}). The roof bolting operation follows immediately after the extraction task and reinforces the roof to provide a safe working environment. Roof bolting is utilized in almost all coal mining operations around the world.

The roof bolt binds the unstable roof together, preventing movement in a rock mass. There are several types of bolt installation techniques, depending on the mechanics of the bolt and the rock.  This application study focuses on a technique where installation of the bolts is done by drilling a hole in the roof, inserting the resin and inserting the bolt. The roof bolting operation is a labor-intensive task that requires the operators of the machine to install and replace detachable drill steels and cutting bits, holding and positioning of resin cartridges and 1.2 to 3 meter (4 to 10 foot) long bolts in a pattern that can be half a meter square. During the roof bolts installation process, the operators are at risk from working in the proximity of potentially unsupported roof, loose bolts, hydraulic-powered equipment, gas and heavy tools in awkward conditions. Apart from these safety risks, the operators are also vulnerable to inhalation of dust and noise from drilling and bolting processes which can be traced to the several pumps from the roof bolter machinery~\cite{jiang2021development}. The operation of these machines requires attention to the risks, which, combined with fatigue, leads to accidents, injuries and severe injuries including fatalities. Therefore, more and more research efforts have been put into developing robot systems that are capable of carrying out the sequence of roof-bolting operations to achieve a high-impact health and safety intervention for roof-bolter operators~\cite{van2013automated,schafrik2022}.

The bolting machines have been automated before, but these modifications are not popular with the community because autonomous machines are highly restricted in their usage.  They are setups for a single-purpose drilling and bolting operation, where in most mining and civil construction, flexible installation is desired.

\begin{figure}[!t]
\centering
\includegraphics[width=0.95\linewidth]{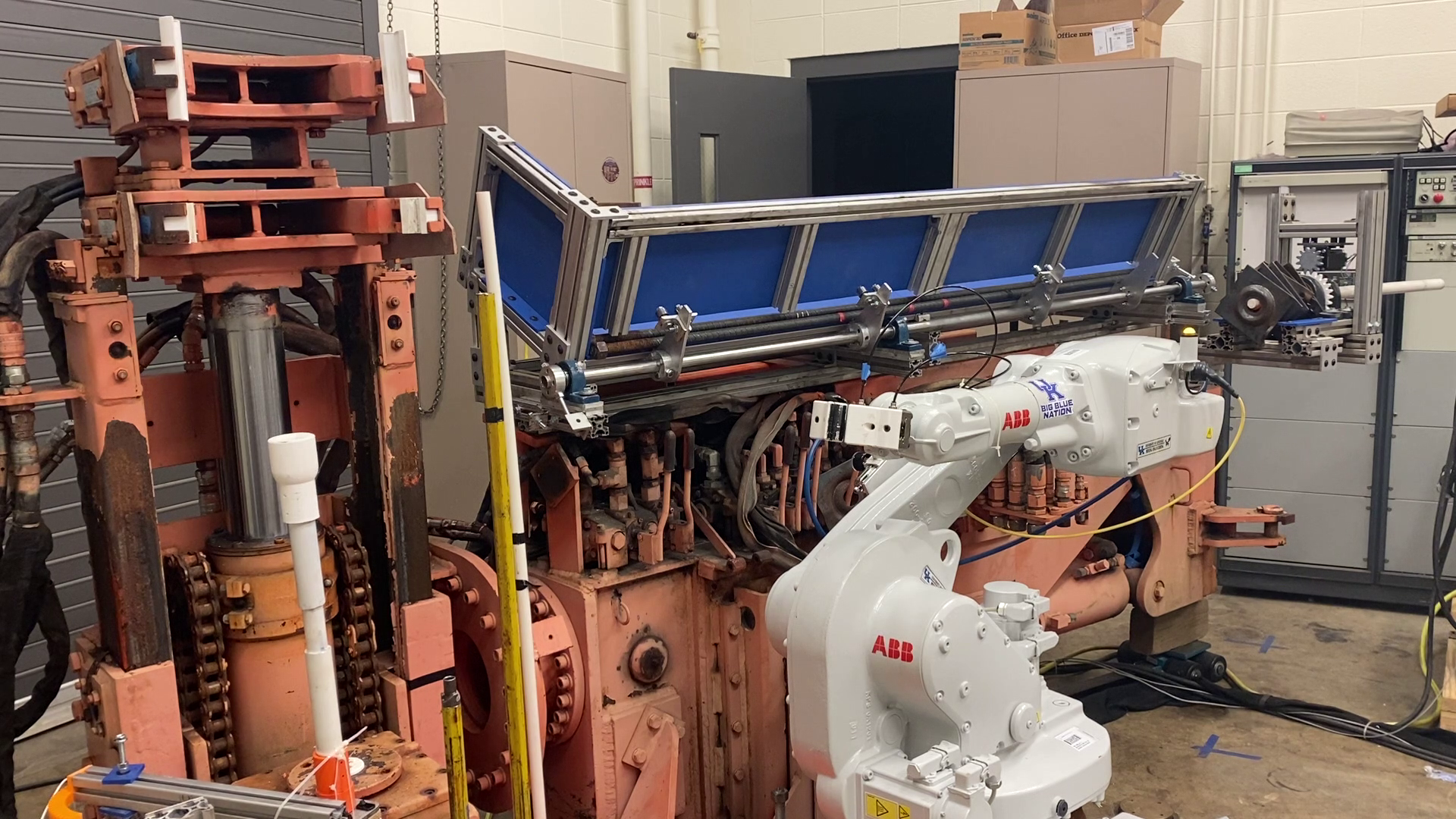}
\caption{The roof bolting system.}
\label{fig:rb}
\end{figure} 

Fig.~\ref{fig:rb} shows a robot-assisted roof-bolting system constructed in our lab~\cite{schafrik2022}. In a roof-bolting task, the system does following actions step-by-step: (\romannum{1}) drill a hole in the roof with a drill steel; (\romannum{2}) remove the drill steel; (\romannum{3}) install resin; (\romannum{4}) install a bolt. To perform these actions successfully, the roof-bolter operator and the roof bolter need to collaborate seamlessly. The role of the roof bolter is to drill the roof and install the resin and the bolt into the roof. The role of the operator is to pick up the drill steel, the resin and the bolt and hand them over to the roof bolter. In our robot-assisted roof-bolting system, we replace the human roof-bolter operator with an ABB IRB 1600 robot because of its high accuracy and flexibility.

Industrial robots have been widely deployed in factories~\cite{nikolaidis2012human} in isolation from people, where their tasks can be pre-defined in the form of waypoints. However, underground mine is usually cluttered and dynamic. For example, human workers who are focused on other tasks may leave tools around unconsciously. The left tools and other objects in the environment will become obstacles blocking the roof-bolting operation. The robot arm then has to clear its operation space, i.e., move movable obstacles out of the way. Moreover, to perform roof-bolting tasks, it is critical to coordinate the roof bolter and robot arm because of their different capabilities. On  one hand, we need the roof bolter to drill holes in the roof and install the bolts into the roof; on the other hand, we need the robot arm to hand bolts, resins and drill steels over to the roof bolter and rearrange movable obstacles. To automatically generate manipulation plans to coordinate the roof bolter and the robot arm, the planning framework should first compute the occlusion and reachability information for the roof bolter and the robot arm (Sec.~\ref{sec:rep}) and then generate effective manipulation plans accordingly.

We observe that in each step of the roof-bolting operation we have a target object whose target configuration is specified and numerous objects that can be treated as movable obstacles. By treating the roof bolter as the second robot, we propose to formulate each step of the roof-bolting operation as a multi-robot geometric task-and-motion planning (MR-GTAMP) problem.

\begin{figure}[!t]
\centering

\captionsetup{justification=centering}

\includegraphics[width=0.8\linewidth]{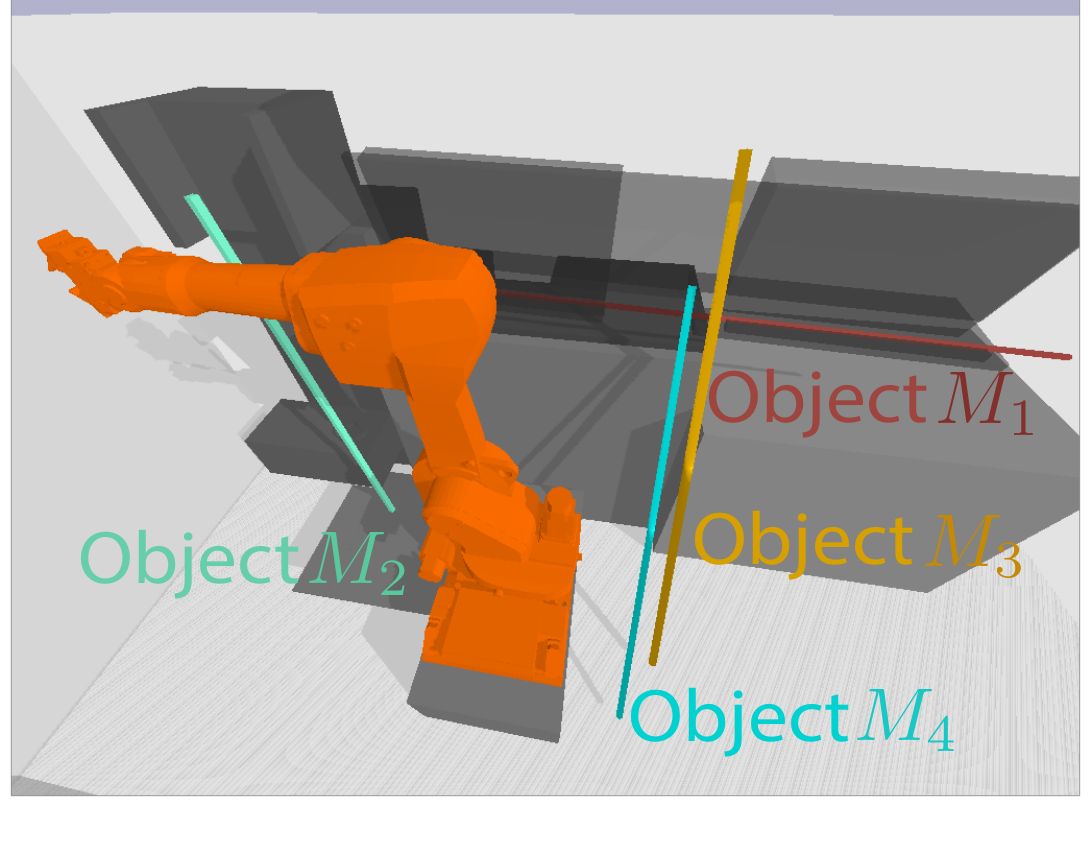}
\caption{Example scenario 1 in the simulation.}
\label{fig:scenario1}
\end{figure}

\begin{figure}[!t]
\centering

\includegraphics[width=0.9\linewidth]{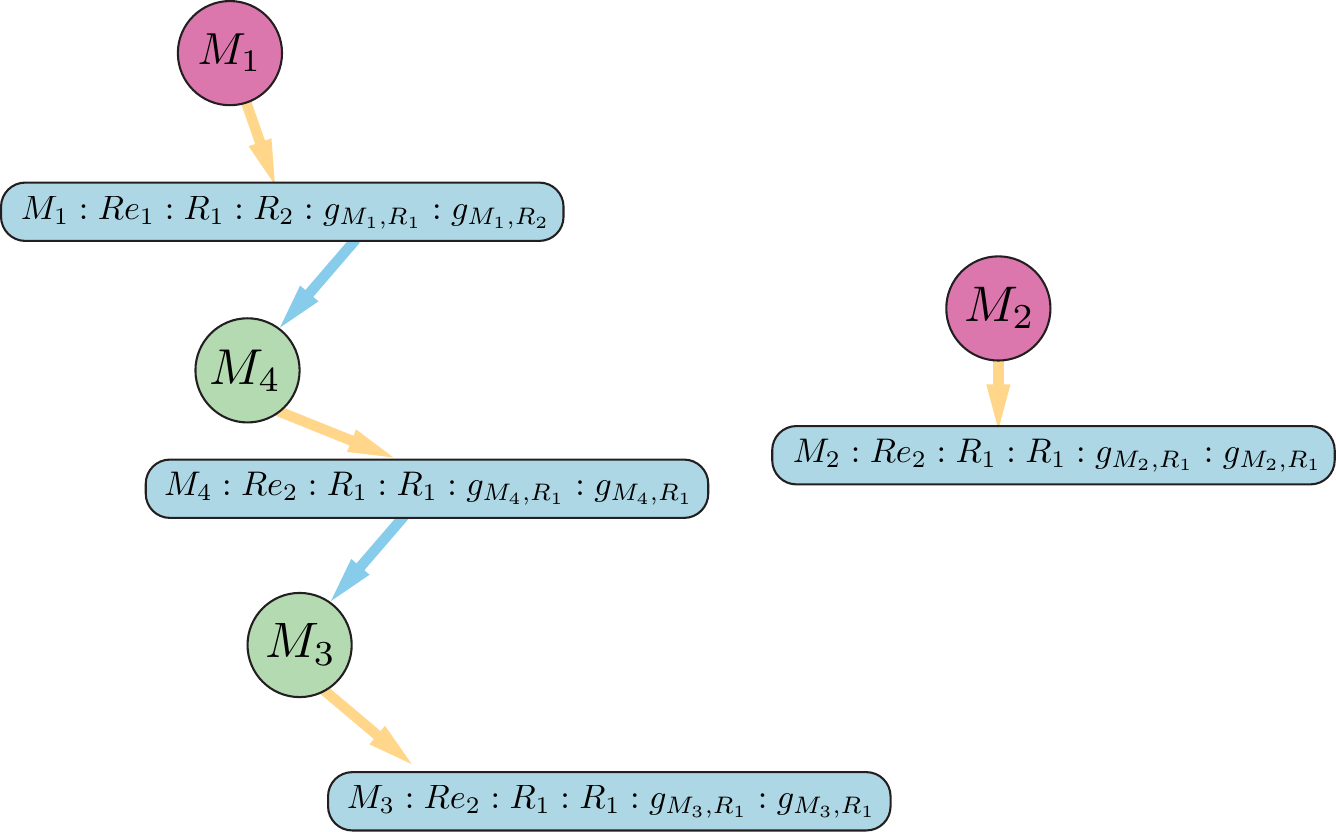}
\caption{Generated CMTGs for example scenario 1. $R_1$ and $R_2$ represent the ABB robot arm and the roof bolter.}
\label{fig:scenario1_cmtg}
\end{figure}

\begin{figure}[!t]
\centering

\captionsetup{justification=centering}

\includegraphics[width=0.8\linewidth]{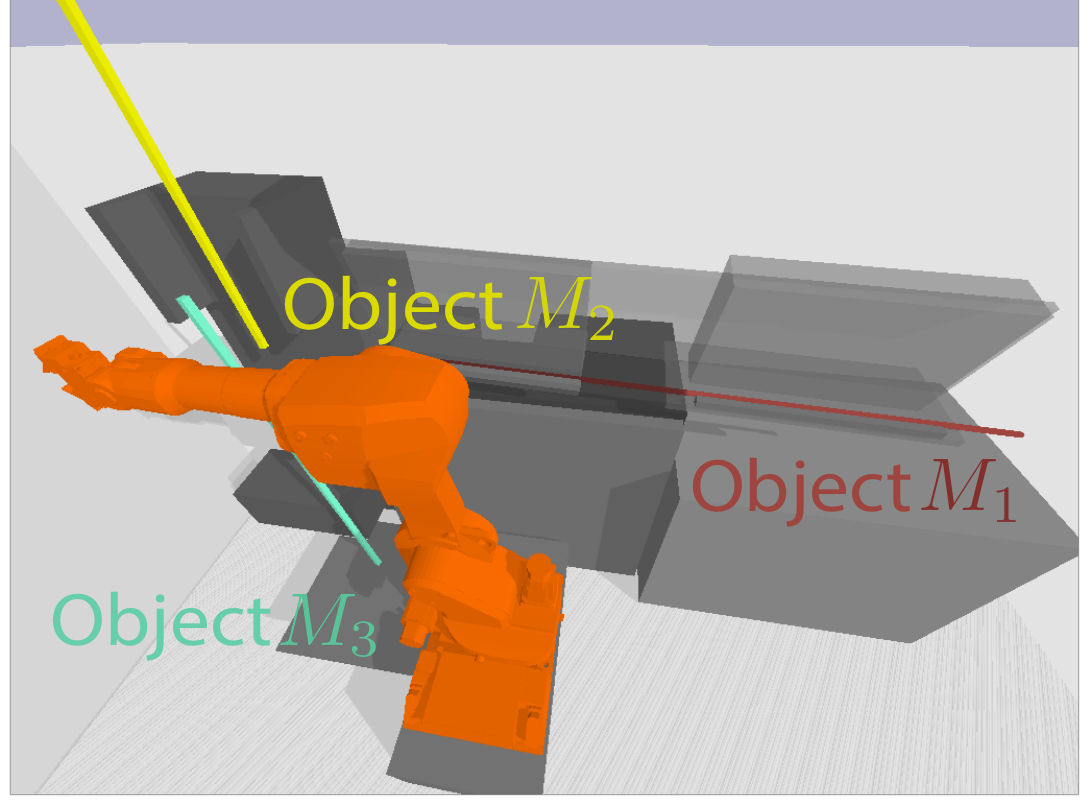}
\caption{Example scenario 2 in the simulation.}
\label{fig:scenario2}
\end{figure}

\begin{figure}[!t]
\centering

\includegraphics[width=0.9\linewidth]{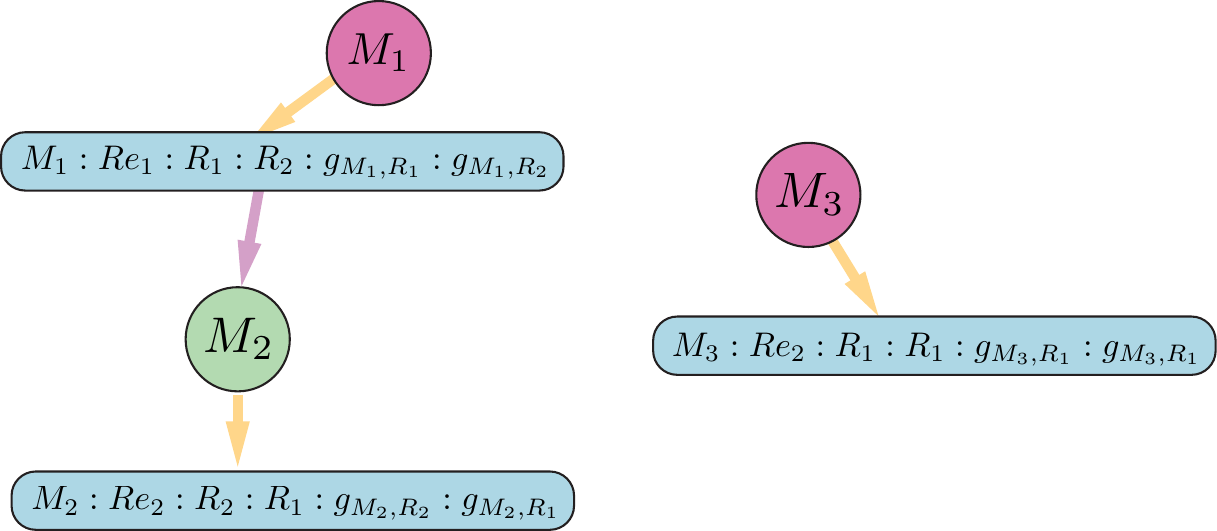}
\caption{Generated CMTGs for example scenario 2. $R_1$ and $R_2$ represent the ABB robot arm and the roof bolter.}
\label{fig:scenario2_cmtg}
\end{figure}

\subsection{Formulating Roof-Bolting Operation as MR-GTAMP problems}

In the roof-bolting task, we need to move the drill steel, the resin and the bolt to their target configurations in the roof. In our application study, we only focus on bolt placement. Other actions can be formulated as MR-GTAMP problems similarly. We assume the target configuration of the bolt has been pre-defined. We formulate an MR-GTAMP problem where we have two robots, i.e., a roof bolter and an ABB IRB 1600 robot arm. These two robots have different reachability: the roof bolter can place the bolt into its target configuration, whereas the robot arm can pick up the bolt from its initial configuration. Moreover, the robot arm can reach most of the movable objects in the environment.

The reachability of the roof bolter and the robot arm can be computed by calling motion planning algorithms (Sec.~\ref{sec:rep}) and can be easily encoded using collaborative manipulation task graphs (CMTGs) (Sec.~\ref{sec:approx}). We will then use our proposed framework to compute executable task-and-motion plans for the roof bolter and the robot arm that account for their different manipulation capabilities. 

\subsection{Two Example Scenarios}

In our application study, we run our proposed planner for two example scenarios. We show the environment setups in simulation and the built CMTGs (Fig.~\ref{fig:scenario1},\ref{fig:scenario1_cmtg},\ref{fig:scenario2},\ref{fig:scenario2_cmtg}). We denote the ABB robot arm and the roof bolter as $R_1$ and $R_2$. For each action, we denote the object that is moved as $M_i$, the grasp that is used by robot $R_k$ as $g_{M_i, R_k}$ and the region to which the object is moved as $Re_j$.

\noindent\textbf{Example scenario 1.} In the first example scenario, we have the bolt as a target object (object $M_1$) and three movable obstacles (objects $M_2$, $M_3$, $M_4$) (Fig.~\ref{fig:scenario1}). The CMTG for moving object $M_1$ is shown in Fig.~\ref{fig:scenario1_cmtg} (left). The CMTG shows that to move object $M_1$, the ABB robot arm and the roof bolter have to perform a handover action. Object $M_4$ blocks the ABB robot arm from picking up object $M_1$ and object $M_3$ blocks the ABB robot arm from picking up object $M_4$. Given the CMTG, we can generate a task skeleton. During grounding (Sec.~\ref{sec:inst}) the generated task skeleton, the planner finds that object $M_2$ blocks the handover action between the ABB robot arm and the roof bolter. The planner then generates a new CMTG to move object $M_2$ (Fig.~\ref{fig:scenario1_cmtg} (right)).

\noindent\textbf{Example scenario 2.} In the second example scenario, we have a target object, bolt (object $M_1$) and two movable obstacles (objects $M_2$, $M_3$) (Fig.~\ref{fig:scenario2}). The CMTG for moving object $M_1$ is shown in Fig.~\ref{fig:scenario2_cmtg} (left). The CMTG shows that to move object $M_1$, the ABB robot arm and the roof bolter have to perform a handover action. Object $M_2$ blocks the roof bolter from placing object $M_1$ to its target configuration. During grounding (Sec.~\ref{sec:inst}) the generated task skeleton based on the CMTG, the planner finds that object $M_3$ blocks the handover action between the ABB robot arm and the roof bolter. The planner then generates a new CMTG to move object $M_3$ (Fig.~\ref{fig:scenario2_cmtg} (right)).

\begin{figure*}[!t]
\centering

\includegraphics[width=0.75\linewidth]{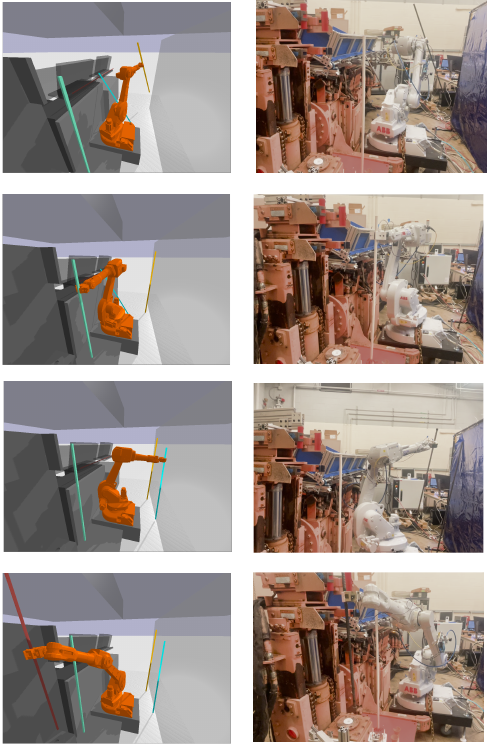}
\caption{Frames showing the execution of the generated plan for example scenario 1 in both simulation (Left) and real-world (Right).}
\label{fig:scenario1_exe}
\end{figure*}

\begin{figure*}[!t]
\centering

\captionsetup{justification=centering}

\includegraphics[width=0.75\linewidth]{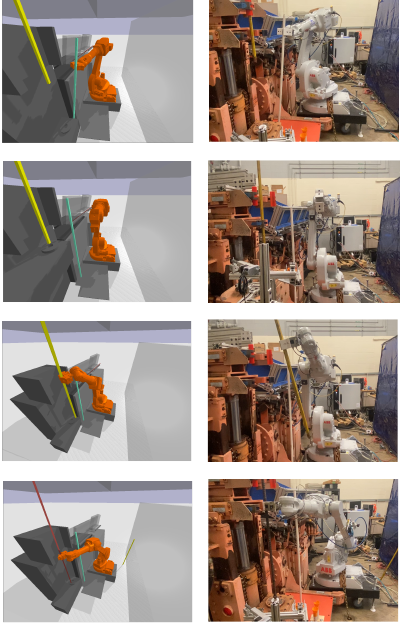}
\caption{Frames showing the execution of the generated plan for example scenario 2 in both simulation (Left) and real-world (Right).}
\label{fig:scenario2_exe}
\end{figure*}

\subsection{Planning and Execution Details}

\noindent\textbf{Planning.} We conduct $5$ trials on an AMD Ryzen Threadripper PRO 3995WX Processor with a memory of 64GB for each scenario. The average planning time for example scenario 1 and example scenario 2 are $144.1 (\pm 21.5)$ seconds and $100.8 (\pm 15.4)$ seconds. We observe that most of the planning time is spent on task skeleton grounding (Sec.~\ref{sec:inst}) where motion planning is extensively called. The average planning time spent on motion planning for example scenario 1 and example scenario 2 are $143.4 (\pm 21.5)$ seconds and $100.2 (\pm 15.4)$ seconds. This is because it is challenging to plan collision-free motion trajectories to move large objects such as the bolt and drill steel in a confined workspace. On the other hand, it only takes $0.6 (\pm 0.0)$ seconds and $0.6 (\pm 0.0)$ seconds on average to compute task skeletons (Sec.~\ref{sec:approx}) for example scenario 1 and example scenario 2. 

\noindent\textbf{Execution.} In Fig.~\ref{fig:scenario1_exe} and Fig.~\ref{fig:scenario2_exe} we show the execution of the generated plans in simulation and real-world. We include videos of scenarios 1 and 2 in the supplemental material. The execution time of the generated plans for example scenario 1 and example scenario 2 are $250.0$ seconds and $270.0$ seconds. To execute the planned motion trajectories on the ABB robot, we first manually smooth the motion trajectories by downsampling the waypoints of the motion trajectories. We then automatically generate ABB robot instructions in RAPID~\cite{robotics2007operating} from the waypoints. Each waypoint is a robot configuration defined in the ABB robot's joint space and is as an argument passed to \textsc{MoveJ} command in RAPID.

\section{\update{Discussion}}

In this paper, we presented a framework for MR-GTAMP problems by proposing a novel MIP formulation to utilize information about the collaborative manipulation capabilities of the individual robots to generate promising task skeletons for guiding the planning search. We proposed an efficient task-skeleton grounding algorithm inspired by the previous work on MAMO~\cite{4209604}. The proposed components are integrated via a Monte-Carlo Tree Search exploration strategy that searches for effective task-and-motion plans. We showed that our framework outperforms two baselines on two challenging MR-GTAMP problems with respect to the planning time and success rates, can generate effective plans with respect to the resulting plan length and the number of objects moved, and can scale up to large problem instances. We also showed that our framework can be applied in the roof-bolting operation for underground mining, where a robotic arm coordinates with an autonomous roof bolter.

\noindent\textbf{\update{Limitations.}} \update{Our work is limited in many ways. In our work, we consider only \textit{monotone} instances of the MR-GTAMP problem, where each object is moved only once. This assumption limits us from solving problem instances that require moving one object multiple times~\cite{pmlr-v100-kim20a} such as Tower of Hanoi, object swapping tasks. We leave the extension to \textit{non-monotone} problem instances for future work. Our framework also pre-defines handover regions for different robots to compute collaborative manipulation information (Sec.~\ref{sec:rep}). This approach may be limited for dynamic environments such as human homes, thus we plan to incorporate a handover region searching process in the task-skeleton grounding component (Sec.~\ref{sec:inst}) in the future. We have also assumed full observability of the scene and therefore cannot handle uncertainties, noise in robot perception~\cite{muguira2022visibility}. We plan to account for sensing limitations in the future~\cite{nikolaidis2009optimal,7451762}. Currently, our approach aims to generate plans with short plan lengths and small numbers of moved objects. However, we do not consider the length of the resulting motion trajectories and the corresponding robot execution time~\cite{chen2022cooperative,marc2023towards}, thus we plan to account for these evaluation metrics in the future.}

Future work also includes using learning to improve the planning efficiency~\cite{doi:10.1177/02783649211038280} and extending the developed techniques to more general MR-TAMP problems~\cite{9636119} and more diverse environments~\cite{FontaineB-RSS-21,Zhang_Fontaine_Hoover_Togelius_Dilkina_Nikolaidis_2020}.

\section{Ethical Statement}
\textbf{Conflict of Interest.} The authors have no competing
interests to declare that are relevant to the content of this
article.

\section{Author Contribution}
H.Z. led the algorithm development, experiments, and paper editing; S.C. and J.L. helped with the algorithm development and paper editing; J.Z. and P.K. helped with the experiments; S.K., Z.A., S.S. and S.N. supervised the project.

\section{Acknowledgements}
This work was supported by the National Science Foundation NRI \# 2024936, the Agilent Early Career Professor Award and the Alpha Foundation for the Improvement of Mine Safety and Health \# AFC820-68.

\small 
\bibliographystyle{sn-mathphys}
\bibliography{sn-bibliography}%

\end{document}